\documentclass[runningheads]{llncs}

\PassOptionsToPackage{table}{xcolor}

\usepackage{eccv}

\usepackage{eccvabbrv}

\usepackage{graphicx}
\usepackage{booktabs}

\usepackage[accsupp]{axessibility}  % Improves PDF readability for those with disabilities.

\usepackage[utf8]{inputenc}
\usepackage[T1]{fontenc}

\usepackage{amsmath, amssymb, amsfonts}
\DeclareMathOperator*{\argmin}{arg\,min}
\usepackage{bm}
\usepackage{nicefrac}

\usepackage{microtype}
\usepackage{textcomp}
\usepackage{soul}
\setul{0.3ex}{0.1ex}

\usepackage{wrapfig}
\usepackage{multirow}
\usepackage{makecell}
\usepackage{threeparttable}

\usepackage{algorithm}
\usepackage{algpseudocode}

\usepackage{enumitem}

\usepackage{hyperref}

\usepackage{orcidlink}

\makeatletter
\expandafter\let\csname showhyphens \endcsname\showhyphens
\makeatother

\begin{document}

% ---------------------------------------------------------------
\title{ToDRE: Effective Visual Token Pruning via Token Diversity and Task Relevance}

% TODO REVIEW: If the paper title is too long for the running head
\titlerunning{ToDRE: Effective Visual Token Pruning}

\author{
Duo Li\inst{1}\textsuperscript{*} \and
Zuhao Yang\inst{1}\textsuperscript{*} \and
Xiaoqin Zhang\inst{2} \and
Ling Shao\inst{3} \and
Shijian Lu\inst{1}\textsuperscript{\textdagger}
}

\authorrunning{D. Li et al.}

\institute{
College of Computing and Data Science, Nanyang Technological University, Singapore\\
\email{\{duo001,yang0756\}@e.ntu.edu.sg}
\and
College of Computer Science and Technology, Zhejiang University of Technology, China
\and
Terminus AI Lab, University of Chinese Academy of Sciences, China
}

\maketitle

\begingroup
\renewcommand{\thefootnote}{%
  \ifcase\value{footnote}%
  \or *%
  \or \textdagger%
  \fi}
\setcounter{footnote}{0}
\footnotetext[1]{Equal contribution.}
\footnotetext[2]{Corresponding author.}
\endgroup

\begin{abstract}
    Visual token pruning, which aims to compress and prune redundant visual tokens, plays a critical role in efficient inference with large vision-language models (LVLMs).
    However, existing methods fail to disentangle intra-modal visual redundancy from cross-modal redundancy between vision and language. We show that visual token diversity and task-specific token relevance are two crucial yet orthogonal factors that complement each other in conveying useful information and should therefore be treated separately for more effective visual token pruning.
    Building upon this insight, we design \textbf{\textsc{ToDRE}}, a two-stage and training-free framework that incorporates \textbf{To}ken \textbf{D}iversity and task \textbf{RE}levance for effective token compression and efficient LVLM inference.
    Instead of pruning redundant tokens, we introduce a greedy max-sum diversification algorithm that selects and retains a subset of diverse and representative visual tokens after the vision encoder.
    On top of that, ToDRE leverages an ``information migration'' mechanism to eliminate task-irrelevant visual tokens within certain decoder layers of the large language model (LLM), further improving token pruning and LVLM inference.
    Extensive experiments show that ToDRE prunes 90\% of visual tokens after the vision encoder as well as all visual tokens in certain LLM decoder layers, leading to a 2.6$\times$ speed-up in total inference time while maintaining 95.0\% model performance plus excellent model compatibility. The code is available at: \url{https://github.com/Yrdal3910/ToDRE/tree/main}. 
\end{abstract}

% Teaser figure
\begin{figure}[t]
    \centering
    \includegraphics[width=\textwidth]{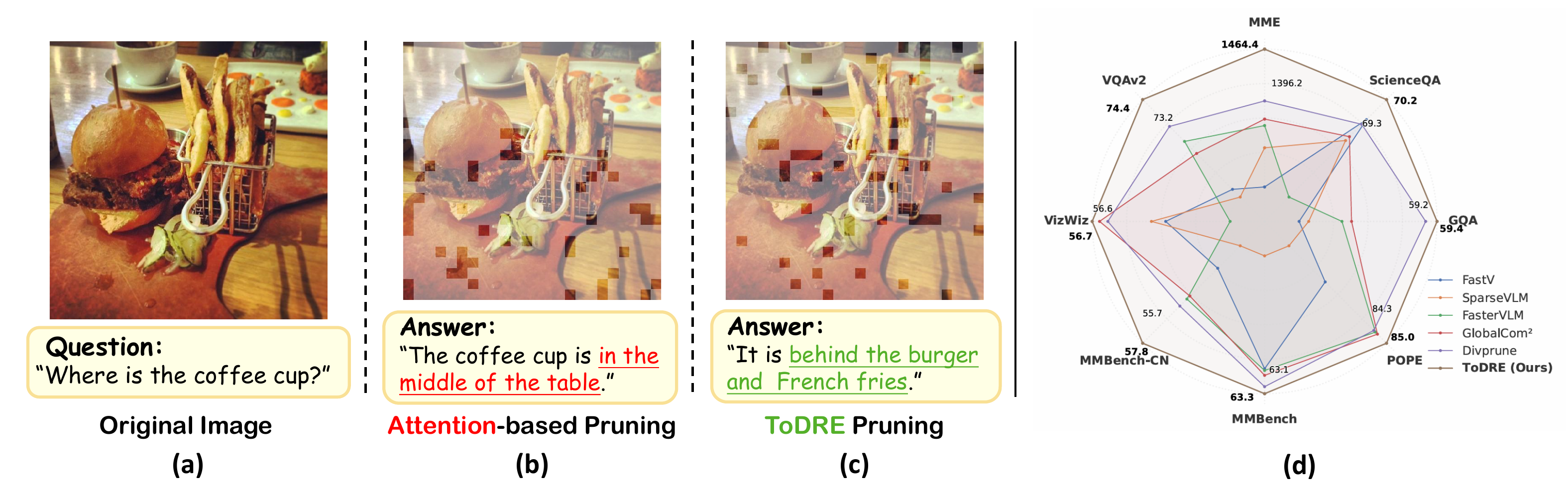}
    \caption{
    \textbf{(a--b)} Existing methods fail to disentangle intra-modal visual redundancy from cross-modal vision-language redundancy, leading to erroneous pruning~\cite{chen2024fastv, zhang2024fastervlm}; as shown in (b), attention-guided pruning discards the informative white coffee-cup region. \textbf{(c)} ToDRE disentangles redundancy into two orthogonal factors, visual token diversity and task-specific token relevance, and prunes them in dedicated stages, preserving critical visual cues. \textbf{(d)} Across eight image-language benchmarks, ToDRE consistently outperforms all baselines.}
    \label{fig:teaser}
\end{figure}

\section{Introduction}
\label{sec:intro}

Leveraging the superior reasoning capability of large language models (LLMs) \cite{achiam2023gpt, bai2023qwen, team2023gemini,touvron2023llama1,touvron2023llama2}, large vision-language models (LVLMs) \cite{bai2025qwen25vl,hurst2024gpt4o,team2025kimivl,wu2024deepseekvl2, zhu2025internvl3} have achieved impressive performance in various multimodal understanding tasks such as visual question answering \cite{goyal2017vqav2,gurari2018vizwiz, hudson2019gqa, lu2022sqa,singh2019textvqa} and video understanding \cite{fu2024videomme, li2024mvbench, mangalam2023egoschema,wu2024longvideobench,zhou2024mlvu}.
LVLMs convert visual inputs into visual tokens and align the converted visual tokens with text tokens for various multimodal understanding tasks.
However, LVLM inference often incurs prohibitive computational and memory costs due to the massive number of visual tokens involved, significantly restricting the applicability of LVLMs to various downstream tasks.

Two representative approaches have recently been explored for improving the LVLM inference efficiency.
The first approach is \emph{model-centric}.
It speeds up the inference via knowledge distillation \cite{cai2024llavakd}, parameter quantization \cite{xie2024quant}, or transformer replacement \cite{qiao2024vlmamba}.
However, this approach requires model retraining, which incurs significant computational costs. The second approach is \emph{data-centric}.
It works by token pruning \cite{chen2024fastv,lin2025vtw, liu2024mustdrop,shang2024prumerge,zhang2024fastervlm} or block skipping \cite{shukor2024skip}, and has attracted increasing attention due to its training-free and architecture-agnostic nature.
Moreover, the \emph{data-centric} approach strikes a strong balance between inference efficiency and model performance, offering a complementary solution to the \emph{model-centric} approach.

\begin{figure}[t]
  \centering
  \includegraphics[width=\linewidth]{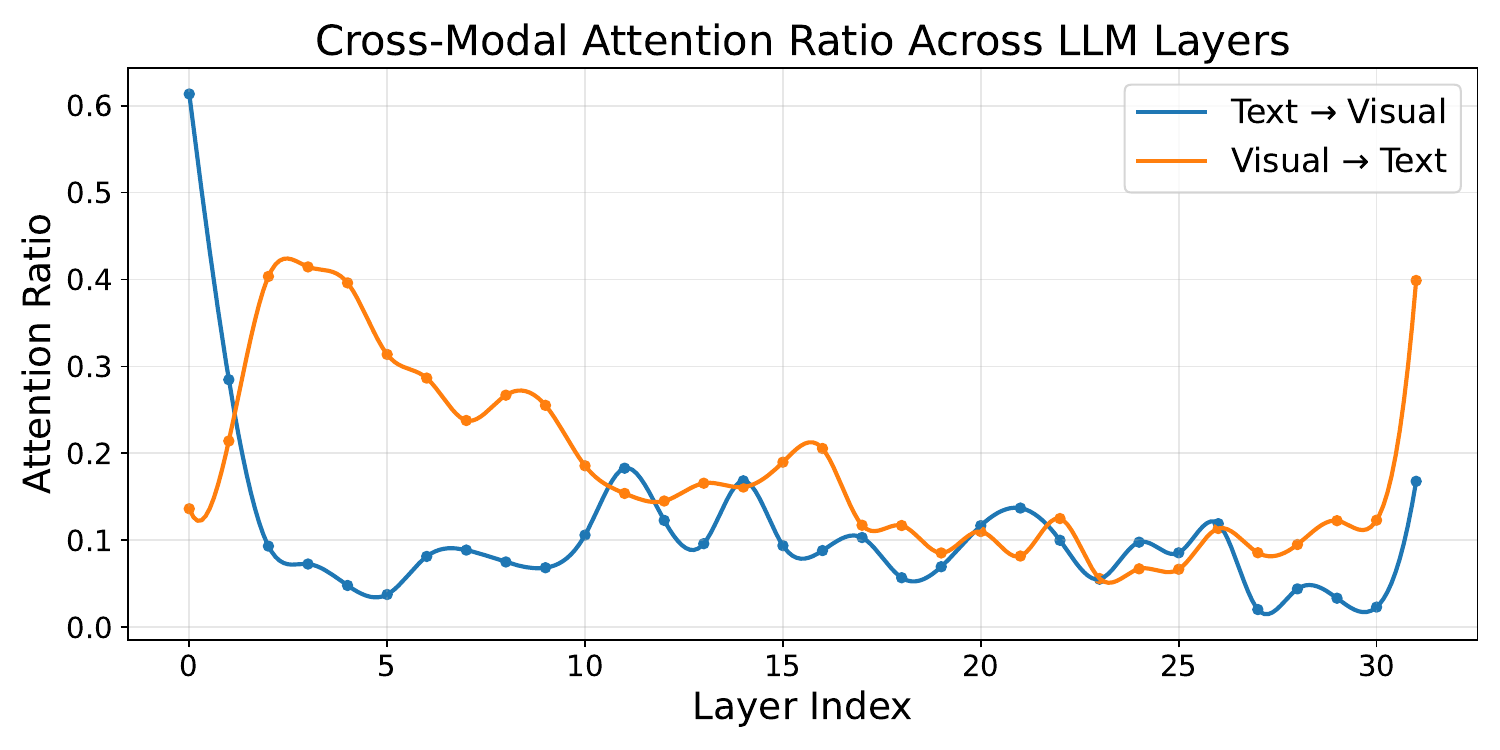}
  \caption{\textbf{Text-to-visual attention (blue) and visual-to-text attention (orange) in each LLM decoder layer.}
  We observe a clear pattern of \emph{``information migration''}: cross-modal attention (both visual-to-text and text-to-visual) is high in early layers, reflecting active information exchange, but gradually diminishes in deeper layers as the model shifts toward unimodal text reasoning.}
  \label{fig:attn_mig}
\end{figure}

Most existing token pruning methods estimate visual redundancy without explicitly differentiating between intra-modal and cross-modal signals. While such metrics\textemdash{}whether based on attention scores \cite{chen2024fastv,zhang2024sparsevlm, shang2024prumerge, zhang2024fastervlm}, feature similarity \cite{bolya2022tome, jiang2025gprune, zhao2024gsearch}, or output divergence \cite{lin2025vtw, ye2025fitprune}\textemdash{}capture certain aspects of redundancy, they inherently conflate two distinct sources: intra-modal redundancy (visually duplicated content) and cross-modal redundancy (tokens irrelevant to the textual query). This coupling is analyzed in detail in \Cref{sec:pre}; here we summarize its practical implications: 
% Treating these entangled signals with a one-size-fits-all criterion leads to suboptimal pruning: 
attention-based methods suffer from positional bias \cite{wen2025dart}, similarity-based merging degrades performance compared to direct pruning \cite{han2024ficoco}, and divergence-based approaches require model-specific calibration \cite{lin2025vtw}. Beyond these issues, we further observe an ``\textit{information migration}'' phenomenon (\Cref{fig:attn_mig}): cross-modal attention (both visual-to-text and text-to-visual) is concentrated in the early decoder layers and progressively diminishes in deeper layers.
% Given that output tokens exhibit near-zero attention to visual tokens during decoding (see \Cref{fig:decoding_attn_ratio}), most existing work 
Given that output tokens exhibit near-zero attention to visual tokens during decoding (see Supplementary Sec.~\ref{sec:supp_neg}), this suggests that once visual information has been absorbed into text representations, the
remaining visual tokens become task-irrelevant and can be safely removed. However, most existing work 
\cite{chen2024fastv,shang2024prumerge,zhang2024fastervlm,zhang2024sparsevlm} passes all remaining visual tokens from the prefilling stage into decoding, thereby incurring unnecessary computations.

We design \textbf{\textsc{ToDRE}}, a simple yet effective token pruning technique that incorporates both \emph{visual token diversity} and \emph{task-specific token relevance} for effective token pruning and efficient LVLM inference.
ToDRE performs token pruning in the embedding space prior to LLM input and during the LLM prefilling stage. First, we introduce a greedy max-sum diversification algorithm that iteratively identifies and preserves visual tokens that have minimal cumulative similarity to the selected tokens.
Such token selection in the LLM embedding space explicitly reduces intra-modal redundancy, thereby preserving a broad spectrum of visual information and enhancing token representativeness under high pruning ratios.
In addition, ToDRE leverages the ``\textit{information migration}'' mechanism by adaptively selecting one layer in the latter half of the LLM decoder (where cross-modal interaction has significantly diminished) and dropping all visual tokens within that layer.
This modality-level pruning removes visual tokens irrelevant to the given question and thus further eliminates redundant computation during inference.
As a result, this relevance-guided pruning enables continuous inference-time efficiency gains as the decoding length increases.
As shown in \Cref{fig:teaser} (c--d), ToDRE's two-stage design enables effective visual token compression while preserving unique visual information and maintaining strong accuracy.

In summary, the major contributions of this work are threefold:
\begin{itemize}
    \item \textbf{Revisit redundancy indicators.}
    First, we re-examine the principles of existing indicators on token redundancy and identify their constraints via systematic and comprehensive analysis. On top of that, we prove that inter-token diversity and token-task relevance are two orthogonal factors, and treating them separately enables more effective token pruning.
    \item \textbf{Propose a training-free and plug-and-play framework.}
    Second, we design a two-stage plug-and-play token pruning technique that is fully compatible with efficient attention operators \cite{dao2022flashattention} without requiring any additional training.
    \item \textbf{Conduct extensive empirical validation.}
    Third, extensive experiments over four widely adopted LVLMs and twelve multimodal benchmarks show that ToDRE achieves superior token pruning consistently.
\end{itemize}

\section{Related Work}
\label{sec:related}

\subsection{Large Vision-Language Models}
Large vision-language models (LVLMs) \cite{bai2025qwen25vl,team2025kimivl,zhu2025internvl3} have demonstrated remarkable advancements by extending the reasoning capabilities of pretrained LLMs \cite{bai2023qwen,touvron2023llama1,touvron2023llama2} to image and video comprehension tasks.
Typically, LVLMs employ a vision encoder to extract visual features, which are subsequently projected into the LLM's embedding space via a visual projector (e.g., Q-Former \cite{li2023blip} or MLP \cite{li2024mini,liu2024llavanext}).
To process real-world high-resolution images, previous LVLMs \cite{bai2023qwenvl,liu2024llava15} resize input images to a fixed resolution, which introduces geometric distortion and degrades fine-grained local details.
To tackle this, subsequent studies adopt dynamic tiling \cite{chen2024far,li2024llavaonevision,liu2024llavanext}, which partitions images into regions and encodes each region independently using a shared vision encoder.
However, dynamic tiling can yield thousands of visual tokens, significantly increasing computational overhead.
This issue becomes even more pressing in video-based LVLMs \cite{bai2025qwen25vl,lin2023videollava}, since processing multiple video frames demands significantly more visual tokens.
These challenges highlight the urgent need for accelerating LVLM inference in resource-constrained real-world environments.

\subsection{Token Compression for LVLMs}
Given that \emph{spatially redundant} visual tokens outnumber \emph{information-dense} text tokens by tens to hundreds of times \cite{marr2010vision}, one natural solution to optimize LVLM inference is visual token compression.
Several early attempts \cite{cai2024matryoshka,li2024tokenpacker,li2024llamavid,yao2024deco} modify model components and introduce additional training costs.
More recently, training-free token compression methods have been widely adopted due to their efficiency and effectiveness.
These methods can be categorized into two main groups:
(1) Token compression in the vision encoder \cite{bolya2022tome,liang2022evit,shang2024prumerge}, the LLM decoder \cite{chen2024fastv,lin2025vtw,zhang2024sparsevlm}, or both \cite{han2024ficoco}:
For example, ToMe \cite{bolya2022tome} reduces tokens in the vision encoding phase by merging redundant tokens via a binary soft-matching algorithm.
Other approaches prune tokens during the LLM decoding stage by evaluating token redundancy through criteria such as attention scores with text tokens \cite{chen2024fastv,zhang2024sparsevlm} or observed divergence with LLM outputs \cite{lin2025vtw,ye2025fitprune}.
Subsequent studies \cite{han2024ficoco,liu2024mustdrop,zhu2024focusllava} perform token compression during both stages to further enhance inference efficiency.
(2) Token compression in the LLM embedding space \cite{alvar2025divprune,zhang2024fastervlm}: A representative example is FasterVLM \cite{zhang2024fastervlm}, which measures token redundancy more accurately through the cross-attention between the \texttt{[CLS]} token and visual tokens.
Unlike previous methods, our proposed ToDRE simultaneously reduces tokens in both the LLM embedding space and the LLM decoder.
Our two-stage approach effectively captures both visual token diversity and token-task relevance\textemdash{}two orthogonal yet critical aspects previously overlooked\textemdash{}achieving superior inference efficiency while maintaining competitive performance.

\section{Preliminary Analysis} \label{sec:pre}

Recently, numerous visual token compression techniques have emerged.
Most approaches \cite{alvar2025divprune,chen2024fastv,lin2025vtw,wen2025dart,zhang2024fastervlm} reduce computational redundancy only within specific stages of the LVLM inference process, lacking a systematic analysis and holistic consideration. To bridge this gap, we provide a deeper analysis organized as follows.
In \Cref{sec:3.1}, we review the processing flow of existing LVLMs, identifying where redundant computation arises.
In \Cref{sec:3.2}, we present a theoretical analysis that justifies disentangling redundancy into intra-modal and cross-modal components, which in turn motivates our two-stage token pruning method. Together, these analyses clarify \textit{where }pruning should be applied and \textit{how} it should be performed.

\subsection{Computational Overhead in LVLM Processing Pipeline} \label{sec:3.1}

% \paragraph{Architecture and Processing Flow.}

% Typically, existing LVLMs consist of three main components: a vision encoder, a vision-language projector, and a LLM decoder.
% Both the encoder and decoder are built upon the Transformer blocks \cite{vaswani2017transformer}.
% Given a visual input $V$, the vision encoder extracts visual features, which are then mapped into a sequence of visual token embeddings $E_v$ by the vision-language projector, aligned with the LLM textual embedding space.
% Then, $E_v$ is concatenated with text embeddings $E_t$ and system prompt embeddings $E_s$ to form the input sequence for LLM.
% During the LLM's prefilling stage, all input tokens interact via self-attention to generate a contextualized representation, denoted as $X = \{\boldsymbol{z}_{s_1}, \dots, \boldsymbol{z}_{s_L}, \boldsymbol{z}_{v_1}, \dots, \boldsymbol{z}_{v_M}, \boldsymbol{z}_{t_1}, \dots, \boldsymbol{z}_{t_N}\}$, where $L$, $M$ and $N$ denote the sequence lengths of system prompt token $\boldsymbol{Z}_{s}$, visual token $\boldsymbol{Z}_{v}$, and text token $\boldsymbol{Z}_{t}$, respectively.
% At each Transformer layer, $X$ is projected into keys and values, which are then stored as KV cache.
% In the subsequent decoding stage, keys and values are computed and added only for newly generated tokens, while previously computed key-value pairs are retrieved from the cache directly.

\paragraph{Computational Cost Analysis.} 

Prior studies \cite{han2024ficoco,liu2024mustdrop,li2026comprehensive} have shown that the dominant contributors to inference cost in LVLMs are the vision-encoding stage, the LLM prefilling stage, and the LLM decoding stage, each of which incurs substantial self-attention and feed-forward network (FFN) computations.
Following previous studies \cite{chen2024fastv,wen2025dart}, we formulate the calculation of floating-point operations (FLOPs) as follows:

\begin{equation}
\text{FLOPs}_{\text{encoding}}
  = \text{FLOPs}_{\text{prefilling}} = T \times \left( 4nd^2 + 2n^2d + 2ndm \right),
\end{equation}

\begin{equation}
\begin{aligned}
\text{FLOPs}_{\text{decoding}} 
  &= T \sum_{t=1}^L \left(4 d^2 + 2d(n + t - 1) + 2dm \right) \\
  &= T \left( 4Ld^2 + 2Ld m + dL(2n + L - 1) \right),
\end{aligned}
\end{equation}

where $T$ is the number of transformer layers; $n$ and $L$ respectively denote the lengths of the input and output sequences; $d$ is the size of the hidden state; and $m$ is the intermediate dimension of the FFN.
We take LLaVA-NeXT-7B \cite{liu2024llavanext}, which employs a CLIP-ViT-Large-Patch14 \cite{radford2021clip} vision encoder and a Vicuna-7B-v1.5 \cite{chiang2023vicuna} LLM decoder, as an example.
The relative ratio of FLOPs (with $n$=3000 and $L$=20) is approximately encoding:prefilling:decoding $\approx$ 1:\textbf{63.6}:0.4.
When scaled to LLaVA-NeXT-13B, the relative ratio shifts to 1:\textbf{121.1}:0.8, indicating that the LLM's prefilling and decoding stages roughly double their share of the total computational cost.
This underscores the importance of pruning visual tokens as early as possible\textemdash{}ideally \emph{prior to} or \emph{during} the LLM prefilling stage\textemdash{}to mitigate the exploding computational burden.

\subsection{Intra- and Cross-Modal Redundancy} \label{sec:3.2}

To justify the disentanglement of redundancy into intra-modal and cross-modal components, we develop the following theoretical analysis.

\medskip
\noindent\textbf{Notation.}
Let the visual token embeddings produced by the vision encoder-projector be
\(
  V=\{v_i\}_{i=1}^{N}\subset\mathbb{R}^{d}
\)
and the text tokens be
\(
  T=\{t_j\}_{j=1}^{M}\subset\mathbb{R}^{d}
\).
We map the two modalities onto mutually orthogonal subspaces of a
shared embedding space:
\[
    \mathcal{V} = \operatorname{Span}(W_V), \qquad
    \mathcal{T} = \operatorname{Span}(W_T).
\]
Visual data (e.g.\ images) encode spatial-texture patterns in pixel grids, whereas textual data (e.g.\ language) convey
semantic information through symbol sequences.
To preserve this intrinsic heterogeneity inside a multimodal model, we apply an embedding scheme based on orthogonal subspace decomposition. The resulting orthogonality constraint is:
\[
  W_V^{\top}W_T = 0 \qquad (\mathcal{V}\;\perp\;\mathcal{T}).
\]

\medskip
\noindent\textbf{Intra-modal redundancy.}
\begin{equation}\label{eq:Dk}
  D_{\kappa}(V)
  = \frac{1}{N^{2}}
    \sum_{i\neq j}
    \kappa\!\bigl(W_V^{\top}v_{i},\,W_V^{\top}v_{j}\bigr),
\end{equation}
where
\(
  \kappa:\mathbb{R}^{d}\times\mathbb{R}^{d}\!\to\!\mathbb{R}_{\ge0}
\)
is any non-negative kernel measuring pairwise similarity.

\medskip
\noindent\textbf{Cross-modal redundancy.}
\begin{equation}\label{eq:Rp}
  R_{\rho}(V,T)
  = \frac{1}{N}
    \sum_{i=1}^{N}
    \rho\!\bigl(W_T^{\top}v_{i},\,T\bigr),
\end{equation}
for a redundancy function
\(
  \rho:\mathbb{R}^{d}\times\mathcal{T}\!\to\!\mathbb{R}_{\ge0}.
\)
Equations \eqref{eq:Dk}--\eqref{eq:Rp} provide the two quantitative metrics that underpin our compression strategy.

\begin{remark}[Limitations of entangled metrics]\label{rem:entangled}
Existing pruning metrics attempt to capture redundancy with a mixed
scoring metric $S(v_i)$, implicitly entangling intra-modal and cross-modal signals. This is inherently sub-optimal: a visual token can be \emph{visually unique yet task-irrelevant}, or \emph{highly relevant yet spatially duplicated}. These two cases are Pareto-incomparable, meaning any single scalar must impose an implicit and often suboptimal trade-off between diversity and relevance, which motivates a principled two-stage design that handles each source of redundancy with a dedicated, optimized criterion.
\end{remark}

\medskip
\begin{lemma}[Subspace decorrelation]\label{lem:indep}
Assume \(\mathcal{V}\perp\mathcal{T}\) and that the projected components
\(W_V^{\top}v\) and \(W_T^{\top}v\) are uncorrelated in expectation.\footnote{This is a modeling assumption that holds exactly under a generative model $v = W_V a + W_T b$ where $a \perp b$, and
approximately when the vision-language projector preserves the heterogeneity of the two modalities. We verify this approximation
empirically and provide a perturbation bound in Remark~\ref{rem:approx}.}
Then for any \(v_i,v_j,v_k\in\mathcal{V}\),
\begin{equation}\label{eq:lemma}
  \mathbb{E}\!\bigl[\kappa(v_i,v_j)\,\rho(v_k,T)\bigr]
  = \mathbb{E}\!\bigl[\kappa(v_i,v_j)\bigr]\,
    \mathbb{E}\!\bigl[\rho(v_k,T)\bigr].
\end{equation}
\end{lemma}

\medskip
\noindent\textbf{Conclusion.}
Under the orthogonality constraint \(W_V^{\top}W_T=0\) and the mild
decorrelation assumption in Lemma~\ref{lem:indep},
\begin{equation}\label{eq:cov}
  \mathrm{Cov}\!\bigl(D_{\kappa}(V),\,R_{\rho}(V,T)\bigr)=0,
\end{equation}
which means the two redundancy measures vary along orthogonal statistical
directions.

\medskip
\noindent\textbf{Proof of \eqref{eq:cov}.}\label{sec:proof}
Set \(X_{ij}=\kappa(v_i,v_j)\) and \(Y_k=\rho(v_k,T)\).
By \cref{eq:Dk,eq:Rp},
\[
  D_{\kappa}(V)=\frac{1}{N^{2}}\!\sum_{i\neq j}X_{ij},
  \qquad
  R_{\rho}(V,T)=\frac{1}{N}\!\sum_{k=1}^{N}Y_k.
\]
Expanding the covariance,
\[
  \mathrm{Cov}(D_{\kappa},R_{\rho})
  = \mathbb{E}[D_{\kappa}R_{\rho}]
    - \mathbb{E}[D_{\kappa}]\mathbb{E}[R_{\rho}].
\]

\noindent\emph{Step 1: substitute definitions.}
\begin{align*}
  \mathbb{E}[D_{\kappa}R_{\rho}]
  &=\mathbb{E}\!\Bigl[
      \Bigl(\tfrac{1}{N^{2}}\!\sum_{i\neq j}
        \kappa(W_V^{\top}v_i,W_V^{\top}v_j)\Bigr)
      \Bigl(\tfrac{1}{N}\!\sum_{k=1}^{N}
        \rho(W_T^{\top}v_k,T)\Bigr)
    \Bigr].
\end{align*}

\noindent\emph{Step 2: apply Lemma \ref{lem:indep}.}
Lemma~\ref{lem:indep} gives
\(
  \mathbb{E}[\kappa\,\rho]
  = \mathbb{E}[\kappa]\,\mathbb{E}[\rho]
\),
so
\[
  \mathbb{E}[D_{\kappa}R_{\rho}]
  = \frac{1}{N^{3}}
    \sum_{i\neq j}\sum_{k=1}^{N}
    \mathbb{E}[\kappa]\;\mathbb{E}[\rho]
  = \mathbb{E}[D_{\kappa}]\,\mathbb{E}[R_{\rho}].
\]

\noindent\emph{Step 3: plug back into the covariance.}
\[
  \mathrm{Cov}(D_{\kappa},R_{\rho})
  = \mathbb{E}[D_{\kappa}]\,\mathbb{E}[R_{\rho}]
    - \mathbb{E}[D_{\kappa}]\,\mathbb{E}[R_{\rho}]
  = 0.\;\;\square
\]

\begin{remark}[Relaxation to approximate orthogonality]\label{rem:approx}
The strict constraint $W_V^\top W_T = 0$ is a simplifying assumption.
In practice, the vision-language projector induces a small but non-zero
coupling $W_V^\top W_T = E$ with $\|E\|_{\mathrm{op}} \leq \epsilon$.
Under this relaxation, the covariance bound becomes
$|\mathrm{Cov}(D_\kappa,\, R_\rho)| = \mathcal{O}(\epsilon)$,
indicating that the two redundancy measures remain approximately
decorrelated as long as the cross-subspace leakage is small.
This is empirically supported by the fact that CLIP-style contrastive
pre-training aligns global semantics (via \texttt{[CLS]} tokens) rather
than patch-level features, leaving the patch-level visual subspace
largely orthogonal to the textual subspace.
\end{remark}

\noindent
Thus, intra-modal redundancy and cross-modal redundancy are \emph{uncorrelated} under the orthogonal subspace model, supporting the effectiveness of the two-stage compression paradigm.

\section{Visual Token Pruning with Token Diversity and Task Relevance %ToDRE: Two-stage Visual Token Pruning via Token Diversity and Task Awareness
}
\label{sec:method}

\begin{figure}[t]
    \centering
    \includegraphics[width=\linewidth]{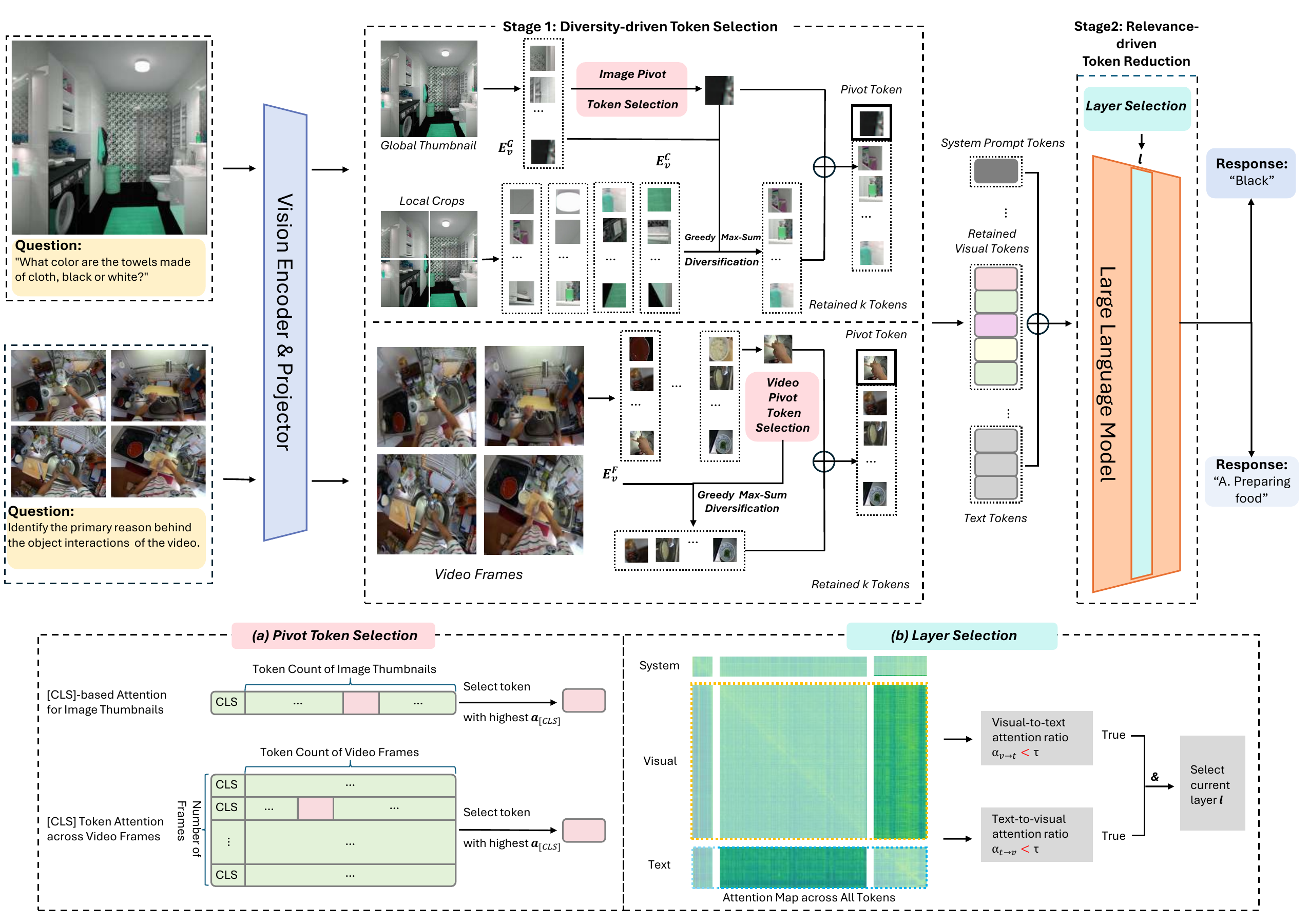}
    \caption{\textbf{Overall framework of ToDRE.} Given the visual and textual inputs, the proposed \textit{Diversity-driven Token Selection} first selects a pivot token from global thumbnail or video frames with \texttt{[CLS]}-based attention and then performs max-sum diversification to retain a diverse set of $k$ visual tokens. The proposed \textit{Relevance-driven Token Reduction} then dynamically identifies a pivot decoder layer and prunes all its visual tokens\textemdash{}the layer is identified if its visual-to-text and text-to-visual attention ratios both fall below a threshold $\tau$. \bm{$E^G_v$}, \bm{$E^C_v$}, and \bm{$E^F_v$} denote the embeddings of thumbnail, local crops, and video frames, respectively.}
    \label{fig:framework}
\end{figure}

Building on the preliminary analysis, we introduce ToDRE, a two-stage, training-free, and plug-and-play visual token compression framework (see \Cref{fig:framework}).
ToDRE utilizes a greedy search in the LLM embedding space to select a maximally diverse subset of visual tokens, followed by an adaptive task-relevance-based pruning mechanism within the LLM decoder. Next, we elaborate on each stage in detail.

\subsection{Diversity-Driven Token Selection %Stage 1: Diversity-driven Token Selection
}
\label{sec:method_stage1}

To obtain a maximally diverse subset of visual tokens, we adopt a greedy max-sum diversification algorithm \cite{borodin2012maxsum} consisting of two steps: (1) initializing a retention set by selecting an initial pivot token, and (2) iteratively adding the token that minimizes its cumulative similarity to the current set.
% Full pseudocode of our proposed token retention algorithm is provided in \Cref{alg:diversity_selection}.
The full pseudocode of our token retention algorithm is provided in
Supplementary Algorithm~\ref{alg:diversity_selection}.

\paragraph{Pivot Token Selection.} 
To determine the initial pivot, we leverage the \texttt{[CLS]} attention from the last layer of the vision encoder \cite{radford2021clip} as an importance indicator.
The attention from the \texttt{[CLS]} token $\boldsymbol{z}_{[\mathrm{CLS}]} \in \mathbb{R}^d$ to other visual tokens $\boldsymbol{Z}_{v} \in \mathbb{R}^{n \times d}$ is calculated as:

\begin{equation}
\begin{aligned}
\boldsymbol{q}_{[\mathrm{CLS}]} &= \boldsymbol{z}_{[\mathrm{CLS}]} \boldsymbol{W}_Q, \quad
\boldsymbol{K}_v = \boldsymbol{Z}_v \boldsymbol{W}_K, \\
\boldsymbol{a}_{[\mathrm{CLS}]} &= 
\operatorname{Softmax}\!\left(\frac{\boldsymbol{q}_{[\mathrm{CLS}]} \boldsymbol{K}_v^{\!\top}}{\sqrt{d}}\right),
\end{aligned}
\end{equation}

where $n$ is the length of the visual token sequence; $d$ is the hidden state size of the vision encoder; $\boldsymbol{W}_Q \in \mathbb{R}^{d \times d}$ and $\boldsymbol{W}_K \in \mathbb{R}^{d \times d}$ represent the weight matrices for queries and keys, respectively.

As shown in \Cref{fig:framework}-(a), pivot token selection proceeds as follows:
(1) \emph{Image Inputs with AnyRes \cite{liu2024llava15} Support}: In this case, LVLM yields one global thumbnail $G$ along with several local crops $C$. We compute the \texttt{[CLS]} attention score for each token in the global thumbnail and choose the token with the highest score as the pivot, since it captures the most comprehensive global information.
(2) \emph{Image Inputs without AnyRes Support}: The pivot token is selected from all visual tokens of the original image, using the same \texttt{[CLS]}-based criterion.
(3) \emph{Video Inputs}: We first identify, for each frame, the visual token with the highest \texttt{[CLS]} attention. The final pivot token is then selected as the one with the highest score among these frame-wise candidates. 

For MLLMs without a \texttt{[CLS]} token in their encoders, a random selection strategy is also acceptable, as it yields performance that is nearly comparable to the original approach. 
% We provide a detailed comparison of different pivot token selection strategies in \Cref{sec:supp_pivot}.
We provide a detailed comparison of different pivot token selection strategies in the Supplementary Sec.\ref{sec:supp_pivot}.

\paragraph{Greedy Max-Sum Diversification.}
The expansion starts from the designated pivot. At iteration $t$, we pick a new token index $c^{(t)}$ by minimizing its \emph{cumulative} similarity to the already selected set:
\begin{equation}
  c^{(t)} \;=\; \argmin_{v \in V \setminus \mathcal{C}^{(t-1)}} 
  \Bigg[ \sum_{c \in \mathcal{C}^{(t-1)}} s(\mathbf{x}_v,\mathbf{x}_c) \Bigg],
  \label{eq:msd_argmin}
\end{equation}
where $\mathbf{x}_v$ and $\mathbf{x}_c$ denote visual token features with indices $v$ and $c$, and $\mathcal{C}^{(t-1)}$ is the selected set from the previous iteration. 
The similarity between two tokens is measured with cosine similarity
\begin{equation}
  s(\mathbf{x}_v,\mathbf{x}_c) \;=\; 
  \frac{\mathbf{x}_v^\top \mathbf{x}_c}{\lVert \mathbf{x}_v \rVert \,\lVert \mathbf{x}_c \rVert}.
\end{equation}
Equivalently, \eqref{eq:msd_argmin} maximizes the \emph{sum of distances} if $d(\cdot,\cdot)=1-s(\cdot,\cdot)$.
After selecting $c^{(t)}$, we update the cumulative similarities by adding its contribution:
\begin{equation}
  \forall v \in V \setminus \mathcal{C}^{(t)}:\quad 
  S_v^{(t)} \;=\; S_v^{(t-1)} \;+\; s(\mathbf{x}_v,\mathbf{x}_{c^{(t)}}),
\end{equation}
and mask the chosen index. This greedy procedure repeats until $k$ diverse tokens (e.g., $k{=}288$, about $10\%$ of visual tokens) are retained, yielding
\begin{equation}
  \mathcal{C} \;=\; \{ c^{(1)}, c^{(2)}, \dots, c^{(k)} \}.
\end{equation}
Finally, all remaining visual tokens are discarded; the retained visual tokens together with all text tokens are fed to the LLM decoder for inference.

\subsection{Relevance-Driven Token Compression}
\label{sec:method_stage2}

While strategies involving partial or multi-stage pruning could be further applied, we argue that such strategies are unnecessary, since the majority of visual tokens have already been removed in the previous stage. We propose a forward-pass metric based on global cross-modal interaction to select the most appropriate decoder layer for token removal. As illustrated in \Cref{fig:framework}-(b), once this layer is identified, we remove all remaining visual tokens immediately after it.

Specifically, let $L$ be the number of decoder layers in the LLM. Based on our observation that cross-modal interaction has largely concluded in the latter half of the decoder (\Cref{fig:attn_mig}), we compute the global cross-modal attention ratios only at a few selected layers in the deeper portion of the LLM to determine whether information migration is complete. Since these attention ratios tend to remain stable across consecutive deeper layers, computing them at every layer would introduce unnecessary overhead. At each selected layer $\ell$, we compute two cross-modal attention ratios based on average attention probabilities across all attention heads and tokens:

\begin{equation}\label{eq:alpha}
\begin{aligned}
\alpha_{t\to v}^{(\ell)} &= 
\frac{\sum_{i\in T}\sum_{j\in V} A_{ij}^{(\ell)}}
     {\sum_{i\in T}\sum_{j\in S\cup V\cup T} A_{ij}^{(\ell)}},\\
\alpha_{v\to t}^{(\ell)} &= 
\frac{\sum_{i\in V}\sum_{j\in T} A_{ij}^{(\ell)}}
     {\sum_{i\in V}\sum_{j\in S\cup V\cup T} A_{ij}^{(\ell)}},
\end{aligned}
\end{equation}

where $A_{ij}^{(\ell)}$ denotes the softmax-normalized attention weight from query token $i$ to key token $j$ at layer $\ell$; $S$, $V$, and $T$ represent the system prompt, visual, and textual tokens, respectively.
If both ratios fall below a threshold $\tau$, indicating that cross-modal interaction has diminished sufficiently, we remove all visual tokens from that layer onward, since the model has already absorbed the necessary visual information into the text representations. 
More detailed ablations of layer selection and the threshold $\tau$ are provided in Supplementary Sec.~\ref{sec:supp_layer} and Sec.~\ref{sec:supp_threshold}, respectively. 
This further eliminates redundant visual computation in the subsequent prefilling and decoding stages, yielding slight improvements in both efficiency and performance.

\section{Experiments}
\label{sec:exp}

\begin{table}[!t]
    \centering
    \resizebox{\linewidth}{!}{
        \begin{tabular}{l|cccccccc|c}
            \toprule[2pt]
            \textbf{Method} & \textbf{MME} & \textbf{ScienceQA} & \textbf{GQA} & \textbf{POPE} & \textbf{MMBench-EN} & \textbf{MMBench-CN} & \textbf{VizWiz} & \textbf{VQAv2} & \textbf{Average} \\
            \midrule
            \multicolumn{1}{@{}l}{} & \multicolumn{8}{c}{\textit{Upper Bound, 2880 Tokens}} & \multicolumn{1}{r@{}}{} \\
            \midrule
            \rowcolor{gray!25}
            LLaVA-NeXT-7B \cite{liu2024llavanext} & 1519.6 & 72.0 & 64.2 & 87.7 & 68.5 & 59.0 & 60.6 & 80.1 & 100.0\% \\
            \midrule
            \multicolumn{1}{@{}l}{} & \multicolumn{8}{c}{\textit{Ratio=25\%, Retain up to 720 Tokens}} & \multicolumn{1}{r@{}}{} \\
            \midrule
            FastV \cite{chen2024fastv} & 1477.3 & 69.8 & 60.4 & 83.1 & 65.6 & 55.4 & 57.2 & 77.2 & 95.4\% \\
            SparseVLM \cite{zhang2024sparsevlm} & 1446.1 & 67.5 & 60.9 & 71.0 & 63.8 & 55.4 & 58.6 & 77.2 & 93.1\% \\
            FasterVLM \cite{zhang2024fastervlm} & 1454.6 & 67.1 & 61.3 & 87.2 & 66.0 & 56.8 & 58.4 & 76.4 & 96.0\% \\
            GlobalCom2 \cite{liu2025globalcom2} & 1468.7 & 68.1 & 61.4 & \textbf{87.6} & 64.0 & 54.4 & 58.7 & 76.6 & 95.6\% \\
            DivPrune \cite{alvar2025divprune} & 1486.5 & 70.0 & 61.8 & 87.4 & 64.6 & 56.4 & 58.5 & 76.4 & 96.6\% \\
            \textbf{ToDRE (Ours)} & \textbf{1504.3} & \textbf{70.7} & \textbf{63.3} & 87.5 & \textbf{66.6} & \textbf{58.0} & \textbf{59.5} & \textbf{77.5} & \textbf{98.2\%} \\
            \midrule
            \multicolumn{1}{@{}l}{} & \multicolumn{8}{c}{\textit{Ratio=10\%, Retain up to 288 Tokens}} & \multicolumn{1}{r@{}}{} \\
            \midrule
            FastV \cite{chen2024fastv} & 1282.9 & 69.3 & 55.9 & 71.7 & 61.6 & 53.5 & 56.1 & 70.2 & 88.8\% \\
            SparseVLM \cite{zhang2024sparsevlm} & 1332.2 & 68.6 & 56.1 & 63.2 & 54.5 & 52.3 & 56.2 & 69.9 & 86.3\% \\
            FasterVLM \cite{zhang2024fastervlm} & 1359.2 & 66.5 & 56.9 & 83.6 & 61.6 & 55.1 & 55.6 & 72.3 & 91.4\% \\
            GlobalCom2 \cite{liu2025globalcom2} & 1365.5 & 68.7 & 57.1 & 83.8 & 61.8 & 55.0 & 56.6 & 71.8 & 92.2\% \\
            DivPrune \cite{alvar2025divprune} & 1396.2 & 69.3 & 59.2 & 84.3 & 63.1 & 55.7 & 56.6 & 73.2 & 93.5\% \\
            \textbf{ToDRE (Ours)} & \textbf{1464.4} & \textbf{70.2} & \textbf{59.4} & \textbf{85.0} & \textbf{63.3} & \textbf{57.8} & \textbf{56.7} & \textbf{74.4} & \textbf{95.0\%} \\
            \midrule[1pt]
            \rowcolor{gray!25}
            LLaVA-NeXT-13B \cite{liu2024llavanext} & 1575.2 & 71.2 & 65.4 & 87.5 & 70.1 & 66.0 & 63.6 & 81.9 & 100.0\% \\
            \midrule
            \multicolumn{1}{@{}l}{} & \multicolumn{8}{c}{\textit{Ratio=25\%, Retain up to 720 Tokens}} & \multicolumn{1}{r@{}}{} \\
            \midrule
            FasterVLM \cite{zhang2024fastervlm} & 1516.1 & 71.1 & 62.3 & 86.1 & 67.6 & 62.1 & 58.1 & 76.1 & 95.6\% \\
            GlobalCom2 \cite{liu2025globalcom2} & 1531.2 & 71.4 & 62.7 & 86.5 & 67.9 & 61.3 & 58.2 & 77.2 & 96.0\% \\
            DivPrune \cite{alvar2025divprune} & 1530.2 & 71.4 & 62.9 & 87.0 & 67.7 & 61.4 & \textbf{60.6} & 77.4 & 96.6\% \\
            \textbf{ToDRE (Ours)} & \textbf{1557.0} & \textbf{72.8} & \textbf{63.8} & \textbf{87.5} & \textbf{69.1} & \textbf{63.9} & 57.6 & \textbf{78.5} & \textbf{97.3\%} \\
            \midrule
            \multicolumn{1}{@{}l}{} & \multicolumn{8}{c}{\textit{Ratio=10\%, Retain up to 288 Tokens}} & \multicolumn{1}{r@{}}{} \\
            \midrule
            FasterVLM \cite{zhang2024fastervlm} & 1386.2 & 70.5 & 58.1 & 81.6 & 61.7 & 53.5 & 55.9 & 77.1 & 89.1\% \\
            GlobalCom2 \cite{liu2025globalcom2} & 1399.5 & 71.0 & 58.3 & 82.4 & 65.0 & 56.6 & 55.6 & 72.8 & 90.8\% \\
            DivPrune \cite{alvar2025divprune} & 1463.3 & 70.7 & \textbf{60.1} & \textbf{86.5} & 64.3 & 53.0 & \textbf{59.1} & 75.4 & 92.5\% \\
            \textbf{ToDRE (Ours)} & \textbf{1490.2} & \textbf{71.4} & 59.9 & 83.7 & \textbf{65.3} & \textbf{60.5} & 56.9 & \textbf{75.9} & \textbf{93.6\%} \\
            \bottomrule[2pt]
        \end{tabular}
    }
    \caption{\textbf{Performance of training-free token compression methods across eight image-language benchmarks.} ``Average'' denotes the mean performance ratio between each token compression method and the vanilla MLLM model. We evaluate all methods at retention ratios of 25\% and 10\%, with the best results highlighted in bold.
}
    \label{tab:image}
\end{table}

\begin{table}[!t]
    \centering
    \resizebox{\linewidth}{!}{
        \begin{tabular}{l|c|c|cccc|c}
            \toprule[2pt]
            \textbf{Method} & \textbf{Retain Ratio} & \textbf{\# Token} & \textbf{VideoMME} & \textbf{Egoschema} & \textbf{MLVU} & \textbf{LongVideoBench} & \textbf{Average} \\
            \midrule
            \rowcolor{gray!25}
            LLaVA-NeXT-7B \cite{liu2024llavanext} & 0\% & 2880 & 33.3 & 35.7 & 20.1 & 42.5 & 100.0\% \\
            \midrule
            FastV \cite{chen2024fastv} & \multirow{4}{*}{25\%} & \multirow{4}{*}{720} & 32.3 & 31.2 & 16.5 & 40.3 & 90.4\% \\
            FasterVLM \cite{zhang2024fastervlm} & & & \textbf{33.8} & 41.0 & \textbf{19.6} & 40.5 & 102.3\% \\
            DivPrune \cite{alvar2025divprune} & & & 33.6 & 41.8 & 19.5 & 40.4 & 102.5\% \\
            \textbf{ToDRE (Ours)} & & & 33.3 & \textbf{42.4} & \textbf{19.6} & \textbf{41.0} & \textbf{103.1\%} \\
            \midrule
            FastV \cite{chen2024fastv} & \multirow{4}{*}{10\%} & \multirow{4}{*}{288} & 30.4 & 30.4 & 11.2 & 38.7 & 80.8\% \\
            FasterVLM \cite{zhang2024fastervlm} & & & \textbf{34.3} & 36.1 & \textbf{19.1} & 36.9 & 96.4\% \\
            DivPrune \cite{alvar2025divprune} & & & 33.3 & 40.9 & 18.9 & 40.3 & 100.7\%\\
            \textbf{ToDRE (Ours)} & & & 33.2 & \textbf{41.9} & 18.2 & \textbf{40.7} & \textbf{100.9\%} \\
            \bottomrule[2pt]
        \end{tabular}
    }
    \caption{\textbf{Performance of training-free token compression methods across four video-language benchmarks.}}
    \label{tab:video}
\end{table}

\begin{table}[t]
\centering
\scriptsize
\setlength{\tabcolsep}{3pt}
\begin{threeparttable}
\resizebox{\columnwidth}{!}{
\begin{tabular}{l c c c c c c}
\toprule[1pt]
& \multicolumn{3}{c}{\textbf{Qwen2.5-VL-7B-Instruct} \cite{bai2025qwen25vl}} & \multicolumn{3}{c}{\textbf{InternVL2-8B} \cite{internvl2}}\\
\cmidrule(lr){2-4}\cmidrule(lr){5-7}
\textbf{Benchmark} & {\textbf{Original}} & {\textbf{Ret.\ 25\%}} & {\textbf{Ret.\ 10\%}} & {\textbf{Original}} & {\textbf{Ret.\ 25\%}} & {\textbf{Ret.\ 10\%}} \\
\midrule
MME \cite{Fu2023mme}         & 1687.7  & 1680.6 & 1573.9  & 1628.8 & 1566.4 & 1424.4   \\
ScienceQA \cite{lu2022sqa}   & 88.5    & 86.4   & 85.4    & 96.5   & 95.3   & 92.5     \\
GQA \cite{hudson2019gqa}         & 60.9    & 57.3   & 52.9    & 62.8   & 56.9   & 52.1     \\
POPE \cite{li2023pope}        & 87.7    & 85.4   & 80.8    & 87.8   & 83.7   & 76.9     \\
MMBench-EN \cite{Liu2023mmbench}  & 82.9    & 79.9   & 72.8    & 81.2   & 77.2   & 71.4     \\
MMBench-CN \cite{Liu2023mmbench}  & 81.7    & 78.0   & 70.4    & 80.0   & 74.2   & 67.6     \\
VizWiz \cite{gurari2018vizwiz}      & 70.6    & 68.7   & 66.7    & 60.6   & 58.6   & 56.3     \\
VQAv2 \cite{goyal2017vqav2}       & 82.9    & 78.3   & 72.8    & 79.0   & 72.8   & 68.6     \\
VideoMME \cite{fu2024videomme}    & 61.5    & 59.4   & 57.3    & 55.0   & 54.2   & 52.6     \\
Egoschema \cite{mangalam2023egoschema}   & 58.3    & 57.3   & 55.8    & 55.9   & 55.0   & 52.5     \\
MLVU \cite{zhou2024mlvu}        & 59.3    & 58.7   & 57.2    & 47.3   & 52.8   & 54.0     \\
LongVideoBench \cite{wu2024longvideobench}  & 58.4  & 57.7   & 54.8    & 55.0   & 52.3   & 48.8     \\
\midrule
\textbf{Average} & 100.0\% & 97.1\% & 92.0\% & 100.0\% & 96.8\% & 91.5\% \\
\bottomrule[1pt]
\end{tabular}
}
\caption{\textbf{Performance of ToDRE on Qwen2.5-VL-7B-Instruct and InternVL2-8B.} Benchmarks as rows. ``Ret.'' = Retention Ratio. Averages are normalized to each model's Original (=100\%).}
\label{tab:qwen-internvl}
\end{threeparttable}
\end{table}

% \begin{table}[t]
% \centering
% \scriptsize
% \setlength{\tabcolsep}{5pt}
% \begin{threeparttable}
% \caption{\textbf{Performance of ToDRE on Qwen2.5-VL-7B-Instruct and InternVL2-8B.} Benchmarks as rows. “Ret.”=Retention Ratio. Averages are normalized to each model's Original (=100\%).}
% \label{tab:retention-all}
% \begin{tabular}{l c c c c c c}
% \toprule
% & \multicolumn{3}{c}{\textbf{Qwen2.5-VL-7B-Instruct} \cite{bai2025qwen25vl}} & \multicolumn{3}{c}{\textbf{InternVL2-8B} \cite{internvl2}}\\
% \cmidrule(lr){2-4}\cmidrule(lr){5-7}
% \textbf{Benchmark} & {\textbf{Original}} & {\textbf{Ret.\ 25\%}} & {\textbf{Ret.\ 10\%}} & {\textbf{Original}} & {\textbf{Ret.\ 25\%}} & {\textbf{Ret.\ 10\%}} \\
% \midrule
% MME \cite{Fu2023mme}         & 1687.7  & 1680.6 & 1573.9  & 1628.8 & 1566.4 & 1424.4   \\
% ScienceQA \cite{lu2022sqa}   & 88.5    & 86.4   & 85.4    & 96.5   & 95.3   & 92.5     \\
% POPE \cite{li2023pope}        & 87.7    & 85.4   & 80.8    & 87.8   & 83.7   & 76.9     \\
% VizWiz \cite{gurari2018vizwiz}      & 70.6    & 68.7   & 66.7    & 60.6   & 58.6   & 56.3     \\
% Egoschema \cite{mangalam2023egoschema}   & 58.3    & 57.3   & 55.8    & 55.9   & 55.0   & 52.5     \\
% MLVU \cite{zhou2024mlvu}        & 59.3    & 58.7   & 57.2    & 47.3   & 52.8   & 54.0     \\
% \midrule
% \textbf{Average} & 100.0\% & 98.2\% & 94.8\% & 100.0\% & 99.5\% & 94.6\% \\
% \bottomrule
% \end{tabular}
% \end{threeparttable}
% \end{table}

\paragraph{Experimental Setting.} We evaluate ToDRE over multiple prevalent LVLMs (including LLaVA-NeXT-7B/13B \cite{liu2024llavanext}, Qwen2.5-VL-7B-Instruct \cite{bai2025qwen25vl}, and InternVL2-8B \cite{internvl2}) and twelve widely adopted benchmarks (including eight on image understanding tasks and four on video understanding tasks). 
% More details on the benchmarks, network backbones, and comparison methods can be found in \Cref{app:exp_setting}. 
More details on the benchmarks, network backbones, and comparison methods can be found in Supplementary Sec.~\ref{app:exp_setting}. 
%Our experiments were conducted using LLaVA-NeXT-7B/13B \cite{liu2024llavanext}, Qwen2.5-VL-7B-Instruct \cite{bai2025qwen25vl}, and InternVL2-8B \cite{internvl2} models on a total of twelve benchmarks, including eight for image understanding and four for video understanding. More details on benchmarks, backbone models and comparison methods can be found in \Cref{app:exp_setting}.

\subsection{Benchmarking %Main Results
}

\paragraph{Image Understanding Tasks.}
In \Cref{tab:image}, we report ToDRE's performance on a range of image-understanding benchmarks at different token-retention ratios.
First, under the same setup where 75\% of visual tokens are pruned in Stage 1\textemdash{}matching competing methods\textemdash{}ToDRE further removes all remaining visual tokens in Stage 2 and achieves a 98.2\% average score, outperforming the second-best method by 1.6\%.
Second, under more extreme compression (only 10\% of visual tokens are retained), ToDRE surpasses the second-best approach by 1.5\%. 
Third, ToDRE also achieves top performance on larger models, reaching an average score of 93.6\% on the 13B variant\textemdash{}demonstrating strong adaptability across model scales. 
Note that FastV \cite{chen2024fastv} and SparseVLM \cite{zhang2024sparsevlm} are excluded from the 13B comparison, as their 7B-tuned pruning strategies degrade substantially when directly transferred to 13B. This further highlights the robustness and transferability of ToDRE.

\paragraph{Video Understanding Tasks.} 
To further assess ToDRE's generalization ability, we evaluate it on both short- and long-form video understanding benchmarks.
As shown in \Cref{tab:video}, ToDRE outperforms the baseline by 3.1\% and 0.9\% under the same token retention ratios used for images, and surpasses the second-best method by 0.6\% and 0.2\%, respectively. Interestingly, ToDRE even surpasses the baseline model in some cases. We attribute this to the reduced interference from redundant visual tokens, which may otherwise suppress task-relevant information during inference.
Similarly, SparseVLM is excluded due to limited transferability, and GlobalCom2 \cite{liu2025globalcom2} is omitted as it targets image-only inputs. In contrast, ToDRE generalizes well across modalities and model scales.

\paragraph{Cross-Model Evaluation.}
As shown in \Cref{tab:qwen-internvl}, we further evaluate ToDRE on Qwen and InternVL backbones. 
Specifically, ToDRE retains 97.1\% and 96.8\% of the original performance on Qwen2.5-VL-7B-Instruct and InternVL2-8B at a 25\% retention ratio, respectively, 
and still maintains more than 90\% of the original performance even when only 10\% of visual tokens are preserved, demonstrating strong robustness across different model architectures.

\subsection{Efficiency}

\begin{table}[!t]
    \centering
    \begin{threeparttable}
    \resizebox{\linewidth}{!}{
        \begin{tabular}{l|ccc|c}
            \toprule[2pt]
            \textbf{Method} & \textbf{\makecell{FLOPs \textdownarrow\\(T)}} & \textbf{\makecell{Memory \textdownarrow\\(GB)}} & \textbf{\makecell{Throughput \textuparrow\\(samples/s)}} & \textbf{Performance \textuparrow} \\
            \midrule
            \multicolumn{1}{@{}l}{} & \multicolumn{3}{c}{\textit{Upper Bound, 2880 Tokens}} & \multicolumn{1}{r@{}}{} \\
            \midrule
            \rowcolor{gray!25}
            LLaVA-NeXT-7B \cite{liu2024llavanext} & 31.4 & 15.9 & 1.5 & 100\% \\
            \midrule
            \multicolumn{1}{@{}l}{} & \multicolumn{3}{c}{\textit{Ratio=10\%, Retain up to 288 Tokens}} & \multicolumn{1}{r@{}}{} \\
            \midrule
            FastV \cite{chen2024fastv}        &  8.2 ($\downarrow$73.9\%) & 14.1 ($\downarrow$11.3\%) & 2.1 (1.4$\times$) & 88.8\% \\
            SparseVLM \cite{zhang2024sparsevlm} &  6.9 ($\downarrow$78.0\%) & 14.1 ($\downarrow$11.3\%) & 2.5 (1.7$\times$) & 86.3\% \\
            FasterVLM \cite{zhang2024fastervlm} &  6.1 ($\downarrow$80.6\%) & \textbf{13.6} ($\downarrow$14.5\%) & 2.7 (1.8$\times$) & 91.4\% \\
            GlobalCom$^2$ \cite{liu2025globalcom2} & 6.1 ($\downarrow$80.6\%) & 13.9 ($\downarrow$12.6\%) & 2.7 (1.8$\times$) & 92.2\% \\
            DivPrune \cite{alvar2025divprune} &  \textbf{6.0} ($\downarrow$80.9\%) & \textbf{13.6} ($\downarrow$14.5\%) & 2.8 (1.9$\times$) & 93.5\% \\
            \textbf{ToDRE (Ours)} &  \textbf{6.0} ($\downarrow$80.9\%) & \textbf{13.6} ($\downarrow$14.5\%) & \textbf{2.9} (1.9$\times$) & \textbf{95.0\%} \\
            \bottomrule[2pt]
        \end{tabular}
    }
    \end{threeparttable}
    \caption{\textbf{Inference efficiency comparisons.} All experiments were conducted on a single NVIDIA RTX 3090 GPU. ``Memory'': peak GPU memory usage; ``Throughput'': number of POPE samples processed per second; ``Performance'': average score across 8 image understanding benchmarks.
    }
    \label{tab:efficiency}
\end{table}

As shown in \Cref{tab:efficiency}, we compare FLOPs, peak memory, throughput, and accuracy across token pruning methods at a fixed 10\% retention ratio. ToDRE achieves the highest throughput on POPE \cite{li2023pope} (2.9 samples/s, 1.9$\times$ over vanilla LLaVA-NeXT-7B) while matching the lowest peak memory (13.6 GB) with FasterVLM and DivPrune \cite{alvar2025divprune}. Despite its efficiency, ToDRE also attains the best average accuracy (95.0\%), exceeding the second-best method by 1.5\%, demonstrating a strong balance among speed, memory, and performance.
In addition, as discussed in \Cref{sec:3.1}, since most benchmarks require only short answers (small $L$), decoding-stage gains are modest. We expect larger benefits for long-form generation where visual tokens otherwise dominate computation.

\subsection{Ablation Study}

\begin{table}[!t]
    \centering
    \resizebox{\linewidth}{!}{
    \begin{tabular}{l|ccccc|c}
        \toprule[2pt]
        \textbf{Method} & \textbf{\makecell{Total Time $\downarrow$ \\ (Min:Sec)}} & \textbf{MME} & \textbf{ScienceQA} & \textbf{GQA} & \textbf{POPE} & \textbf{Average} \\
        \midrule
        \multicolumn{1}{@{}l}{} & \multicolumn{5}{c}{\textit{Upper Bound, 2880 Tokens}} & \multicolumn{1}{r@{}}{} \\
        \midrule
        \rowcolor{gray!25}
        LLaVA-NeXT-7B \cite{liu2024llavanext} & 77:04 & 1519.6 & 72.0 & 64.2 & 87.7 & 100.0\% \\
        Stage 2 only & 70:15 & 1522.7 & 71.9 & 64.3 & 87.6 & 100.0\% \\
        \midrule
        \multicolumn{1}{@{}l}{} & \multicolumn{5}{c}
        {\textit{Ratio=25\%, Retain up to 720 Tokens}} & \multicolumn{1}{r@{}}{} \\
        \midrule
        Stage 1 only & 48:10 & 1503.8 & 70.6 & 63.1 & \textbf{87.5} & 98.8\% \\
        \textbf{Stage 1 + Stage 2 (ToDRE)} & \textbf{44:18} & \textbf{1504.3} & \textbf{70.7} & \textbf{63.3} & \textbf{87.5} & \textbf{98.9\%} \\
        \midrule
        \multicolumn{1}{@{}l}{} & \multicolumn{5}{c}{\textit{Ratio=10\%, Retain up to 288 Tokens}} & \multicolumn{1}{r@{}}{} \\
        \midrule
        Stage 1 only & 31:18 & 1458.6 & 70.4 & \textbf{59.4} & \textbf{85.0} & 95.8\% \\
        \textbf{Stage 1 + Stage 2 (ToDRE)} & \textbf{29:43} & \textbf{1469.3} & \textbf{70.5} & \textbf{59.4} & \textbf{85.0} & \textbf{96.0\%} \\
        \bottomrule[2pt]
        \end{tabular}
    }
    \caption{\textbf{Ablation study on two-stage token compression.} We evaluated the individual and combined effects of the proposed two-stage pruning pipeline under retention ratios of 25\% and 10\%.
}
    \label{tab:ablation_2stage}

\end{table}

We conduct ablation studies to evaluate individual and combined contributions of the two stages in our framework. As shown in \Cref{tab:ablation_2stage}, applying Stage 2 only already reduces the overall inference time by 8.8\% on LLaVA-NeXT-7B baseline (from 77:04 to 70:15), while maintaining a lossless average performance of 100.0\%. The limited efficiency gain is expected, as Stage 2 only accelerates the latter part of inference, and most tasks involve generating very short outputs.

In contrast, applying Stage 1 only yields substantial time savings of 37.5\% (48:10) and 59.4\% (31:18), respectively, with minimal drops in performance. When incorporating both stages (Stage 1 + Stage 2), we observe consistent improvements:
First, at the 25\% ratio, performance improves from 98.8\% to 98.9\% with total time reduced (from 48:10 to 44:18).
Second, at the 10\% ratio, performance increases from 95.8\% to 96.0\%, with total time reduced (from 31:18 to 29:43).
Overall, ToDRE reduces inference time by 42.5\% and 61.4\% at the 25\% and 10\% token retention ratios, respectively, while even improving performance (up to +0.2\% gain). These results show that Stage 2 complements the diversity-based Stage 1, improving the accuracy-efficiency trade-off across compression settings.

\section{Conclusion}
\label{sec:conclusion}

In this work, we systematically analyze redundancy in LVLM inference and identify two key inefficiencies:
(1) redundant visual tokens that inflate intra-modal computation, and
(2) tokens that contribute little cross-modal information during decoding.
To address these inefficiencies, we propose \textsc{ToDRE}, a training-free, architecture-agnostic framework that first selects a maximally diverse subset of visual tokens via a greedy max-sum diversification algorithm, then removes all remaining visual tokens once cross-modal attention fades.
Experiments on twelve image- and video-language benchmarks show that ToDRE prunes up to 90\% of visual tokens while preserving 95.0\% of the original performance, achieving 2.6$\times$ faster inference and 14.5\% lower memory usage than uncompressed baselines.

\section*{Acknowledgment}
\begin{flushleft}
This study is funded by the Ministry of Education, Singapore, under the Tier-2 project scheme with project number MOE-T2EP20123-0003.
\end{flushleft}

\begin{flushleft}
We thank Yichao Gao for helpful discussions and valuable suggestions on the mathematical derivations in this paper.
\end{flushleft}

\bibliographystyle{splncs04}
\bibliography{main}

\appendix

\clearpage
\begin{center}
{\Large\bfseries Supplementary Material\par}
\end{center}
\vspace{1em}
\setcounter{table}{0}
\setcounter{figure}{0}
\renewcommand{\thetable}{A\arabic{table}}
\renewcommand{\thefigure}{A\arabic{figure}}
In this supplementary material, we first provide detailed descriptions of the benchmarks used in our experiments in Sec.~\ref{app:exp_setting}.
Additional ablation studies are presented in Sec.~\ref{sec:supp_ablation}, where we further analyze different pivot token selection strategies in Sec.~\ref{sec:supp_pivot}, as well as the effects of the pruning threshold and layer selection in Sec.~\ref{sec:supp_threshold} and Sec.~\ref{sec:supp_layer}, respectively.
We further investigate the negligible decoding-stage attention to visual tokens in Sec.~\ref{sec:supp_neg}.
In Sec.~\ref{sec:supp_longqa}, we demonstrate the advantages of \textbf{\textsc{ToDRE}} using chatbot-style question-answering cases that require long, free-form responses.
Finally, in Sec.~\ref{sec:supp_visualization}, we provide qualitative comparisons on seven benchmarks to visualize the differences between our two-stage approach and existing \emph{attention-driven} token pruning methods.
Unless otherwise stated, all experiments were conducted on 4 NVIDIA RTX 3090 GPUs.

\begin{algorithm}[htbp]
\caption{Proposed Greedy Max-Sum Diversification for Token Retention}
\label{alg:diversity_selection}
\textbf{Input:} $V \in \mathbb{R}^{n \times d}$: visual features; $\alpha \in \mathbb{R}^n$ or $\mathbb{R}^{f \times t}$: \texttt{[CLS]}-to-token attention; $k$: number of tokens to retain \\
\textbf{Output:} $\mathcal{C}$: indices of selected tokens
\begin{algorithmic}[1]
\Statex \textit{Stage 1: Pivot Token Selection}
\If{$\alpha \in \mathbb{R}^n$} \Comment{Image input}
    \State $p \gets \arg\max \alpha$
\ElsIf{$\alpha \in \mathbb{R}^{f \times t}$} \Comment{Video input: $f$ frames, $t$ tokens per frame}
    \State $(a,b) \gets \arg\max_{1 \leq a \leq f,\,1 \leq b \leq t} \alpha_{a,b}$; \quad $p \gets (a-1)t + b$
\EndIf

\Statex \textit{Stage 2: Greedy Max-Sum Diversification}
\State $X \gets \mathrm{row\_normalize}(V)$ \Comment{$\ell_2$-normalize rows for cosine similarity}
\State $\mathcal{C} \gets \{p\}$
\State $s \gets X X_p^\top$ \Comment{$s_i=\cos(x_i,x_p)$ is cumulative similarity}
\State $s_p \gets +\infty$ \Comment{mask selected index}
\For{$i = 1$ to $k-1$}
    \State $c \gets \arg\min s$ \Comment{pick token with smallest cumulative similarity}
    \State $\mathcal{C} \gets \mathcal{C} \cup \{c\}$
    \State $s \gets s + X X_c^\top$ \Comment{update: add similarity to the new token}
    \State $s_c \gets +\infty$ \Comment{mask the newly selected index}
\EndFor
\State \Return $\mathcal{C}$
\end{algorithmic}
\end{algorithm}

\section{Experimental Details} 
\label{app:exp_setting}

\subsection{Benchmarks}

We evaluate our method on a range of widely used benchmarks that collectively cover various aspects of multimodal intelligence.
For image understanding tasks, we conduct experiments on eight benchmarks: MME \cite{Fu2023mme}, ScienceQA \cite{lu2022sqa}, GQA \cite{hudson2019gqa}, POPE \cite{li2023pope}, MMBench and MMBench-CN \cite{Liu2023mmbench}, VizWiz \cite{gurari2018vizwiz}, and VQAv2 \cite{goyal2017vqav2}.
For video understanding tasks, we evaluate our method on four benchmarks: VideoMME \cite{fu2024videomme}, EgoSchema \cite{mangalam2023egoschema}, MLVU \cite{zhou2024mlvu}, and LongVideoBench \cite{wu2024longvideobench}.

\paragraph{MME.}
MME is a comprehensive benchmark designed to evaluate the perceptual and cognitive capabilities of multimodal models across 14 diverse subtasks. It includes both perception-oriented tasks\textemdash{}such as OCR, object counting, spatial localization, and color recognition\textemdash{}and fine-grained recognition of posters, celebrities, scenes, landmarks, and artworks. All tasks are framed as binary judgment questions, using carefully crafted instruction-answer pairs to reduce data leakage and ensure fairness. We follow the standard protocol and report the perception score for evaluation, based on 2,374 image-question pairs.

\paragraph{ScienceQA.}
ScienceQA is a multimodal benchmark designed to assess a model's zero-shot generalization and reasoning capabilities in scientific domains. It spans three major subject areas\textemdash{}natural science, language science, and social science\textemdash{}with questions hierarchically organized into 26 topics, 127 categories, and 379 skills. The benchmark consists of multiple-choice questions, some accompanied by illustrative images.
In our experiments, we evaluate on the ScienceQA split used by our benchmark protocol, which contains 6,258 question-answer pairs.

\paragraph{GQA.} 
GQA is a benchmark designed to evaluate a model's structured understanding and reasoning capabilities over visual scenes. It is built upon three key components: images, scene graphs, and carefully constructed questions. Each image is accompanied by a scene graph derived from the Visual Genome dataset \cite{Krishna2016visualgc}, which provides detailed object-level annotations, attributes, and relationships within the scene. The questions are automatically generated from these graphs to ensure semantic clarity and logical consistency, enabling fine-grained assessment of a model's reasoning ability. Following standard practice, we report accuracy on the test-dev set, which contains 12,578 image-question pairs.

\paragraph{POPE.}
POPE is a benchmark designed to assess object hallucination in vision-language models through binary questions about the presence of specific objects in images. The images are sourced from the MSCOCO dataset \cite{Lin2014microsoftcc}, and evaluation is based on the average F1 score across three sampling strategies, using a total of 8,910 image-question pairs.

\paragraph{MMBench.}
MMBench is a hierarchical benchmark designed to comprehensively evaluate multimodal model capabilities across three levels: perception and reasoning (L1), six sub-skills (L2), and 20 specific tasks (L3). Each task consists of multiple-choice questions. The benchmark is available in both English and Chinese, with 4,377 and 4,329 image-question pairs, respectively. We use both MMBench and MMBench-CN for evaluation.

\paragraph{VizWiz.}
VizWiz is a real-world benchmark that assesses visual understanding using images captured by blind users, each paired with a natural question. Due to the real-life conditions under which the images are captured, such as motion blur or poor lighting, some questions may be difficult or even impossible to answer. Each question is annotated with 10 crowd-sourced answers for evaluation. We evaluate on the test-dev set, which contains 8,000 image-question pairs.

\paragraph{VQAv2.}
VQAv2 is a benchmark designed to evaluate a model's visual recognition and reasoning capabilities through open-ended questions grounded in real-world images. It contains 265,016 images from the MSCOCO dataset \cite{Lin2014microsoftcc}, with each image paired with at least three questions. To mitigate bias, the dataset adopts an adversarially balanced design, ensuring that each question appears with multiple images leading to different answers. Each question is annotated with ten human-provided answers. We use the test-dev set for evaluation, which includes 107,394 image-question pairs, with scoring based on standard automatic metrics.

\paragraph{VideoMME.}
VideoMME is a comprehensive benchmark designed to evaluate the video understanding capabilities of LVLMs. It comprises 900 videos totaling approximately 254 hours, spanning six primary domains and 30 subcategories. The videos vary in length\textemdash{}short (<2 minutes), medium (4--15 minutes), and long (30--60 minutes)\textemdash{}to assess models across different temporal contexts. Each video is accompanied by three expert-annotated multiple-choice questions. We conduct our evaluation on the full VideoMME dataset, which contains 2,700 video-question pairs.

\paragraph{EgoSchema.}
EgoSchema is a diagnostic benchmark designed to evaluate the very long-form video-language understanding capabilities of LVLMs. Derived from the Ego4D dataset \cite{Grauman2021ego4dat}, it comprises over 5,000 human-curated multiple-choice question-answer pairs spanning more than 250 hours of egocentric video footage, covering a broad range of natural human activities and behaviors. Each question is based on a three-minute-long video clip and requires selecting the correct answer from five options. In our experiments, we evaluate on the EgoSchema test set, which contains 5,031 video-question pairs.

\paragraph{MLVU.}
MLVU is a comprehensive benchmark designed to evaluate the long video understanding capabilities of LVLMs. It comprises a diverse set of videos ranging from 3 minutes to 2 hours in length, with an average duration of approximately 12 minutes. The dataset encompasses various video genres, including movies, documentaries, surveillance footage, egocentric recordings, cartoons, and gameplays, to reflect a wide array of real-world scenarios. We conduct our evaluation on the test-dev set, which contains 2,174 video-question pairs.

\paragraph{LongVideoBench.}
LongVideoBench is a comprehensive benchmark designed to evaluate the long-context video-language understanding capabilities of LVLMs. It comprises 3,763 web-collected videos, each accompanied by subtitles, spanning diverse themes such as movies, news, lifestyle, and educational content. The videos vary in length, ranging from a few seconds up to an hour, to assess models across different temporal contexts. Following standard practice, we report accuracy on the test-dev set, which contains 1,337 video-question pairs.

\subsection{Backbone Models}

\paragraph{LLaVA-NeXT.}
LLaVA-NeXT \cite{liu2024llavanext}, also referred to as LLaVA-1.6, extends LLaVA-1.5 \cite{liu2024llava15} by introducing a dynamic high-resolution processing strategy that enhances performance on tasks requiring visual reasoning, OCR, and world knowledge. In contrast to the fixed-resolution scaling used in LLaVA-1.5, LLaVA-NeXT adaptively adjusts the input resolution by selecting an optimal aspect ratio based on the original image. The resolution can be increased by up to $4\times$. Notably, this enhancement is achieved without modifying the visual encoder. Instead, each high-resolution image is divided into multiple sub-images of the original size, which are independently encoded and then concatenated before being passed to the language model. All experiments in this study are based on the 7B and 13B versions of LLaVA-NeXT. 

\paragraph{Qwen2.5-VL.}
Qwen2.5-VL \cite{bai2025qwen25vl} is the flagship vision-language model in the Qwen family, featuring significant improvements in visual reasoning, localization, document understanding, and long-video comprehension. 
It supports object localization via bounding boxes or points and can output structured data (e.g.\ JSON) from documents, forms, tables, and diagrams.  
To handle complex visual inputs, Qwen-2.5-VL employs dynamic resolution processing and absolute time encoding, which allow it to process variable-resolution images and long-range videos without conventional resizing or normalization.  
A native dynamic-resolution ViT architecture with windowed attention is trained from scratch to balance efficiency and perceptual fidelity.  
In this work, we use the 7B instruction-tuned variant, Qwen-2.5-VL-7B-Instruct, for experiments.

\paragraph{InternVL2.}
The InternVL2~\cite{internvl2} series provides a family of multimodal large language models (MLLMs) available in multiple sizes (e.g., 1B--8B--76B+) and instruction-tuned variants. 
It is trained with long-context modeling to support not only single-image inputs but also multi-image and video comprehension. 
InternVL2 offers broad capability coverage, including document, chart, and OCR understanding, visual reasoning, grounding, and multi-image or video comprehension, while maintaining a consistent architecture across different model scales. 
In our experiments, we employ the 8B instruction-tuned model, InternVL2-8B.

\subsection{Comparison Methods}

We compare our method with a range of representative training-free visual token compression methods, each employing distinct strategies such as attention-guided pruning and adaptive token allocation.

\paragraph{FastV.}
FastV \cite{chen2024fastv} is a training-free method that reduces computational overhead in vision-language models by performing early-stage visual token pruning. It identifies and removes the least relevant tokens after the second LLM layer by averaging attention scores.

\paragraph{SparseVLM.}
SparseVLM \cite{zhang2024sparsevlm} ranks the importance of both visual and textual tokens based on cross-modal attention, and introduces adaptive sparsity ratios along with a token recycling strategy to better utilize discarded information.

\paragraph{VisPruner.}
VisPruner \cite{zhang2024fastervlm} leverages attention from the \texttt{[CLS]} token to visual tokens as an importance indicator, pruning the less relevant visual tokens accordingly.

\paragraph{GlobalCom$^2$.}
GlobalCom$^2$ \cite{liu2025globalcom2} is designed for high-resolution image understanding tasks that receive both a global thumbnail and multiple local crops. The thumbnail provides global contextual guidance for the compression of each crop in a task-specific manner.

\paragraph{DivPrune.}
DivPrune \cite{alvar2025divprune} formulates visual token retention as a \emph{min--max diversity} problem, employing a greedy algorithm that iteratively selects tokens most dissimilar to those already chosen.

\section{More Ablation Studies} \label{sec:supp_ablation}

\subsection{Pivot Token Selection Strategies}
\label{sec:supp_pivot}

\begin{table}[t]
\centering
\scriptsize
\begin{threeparttable}
\begin{tabular}{lccccc}
\toprule
\textbf{Method} & \textbf{MME} & \textbf{ScienceQA} & \textbf{GQA} & \textbf{POPE} & \textbf{Average} \\
\midrule
Original & 1519.6 & 72.0 & 64.2 & 87.7 & 100\% \\
\midrule
\multicolumn{6}{l}{\textit{Retention Ratio = 25\%}} \\
\quad \texttt{[CLS]}   & \textbf{1504.3} & \textbf{70.7} & \textbf{63.3} & \textbf{87.5} & \textbf{98.9\%} \\
\quad Random    & 1500.0 & 70.6 & 63.2 & 87.1 & 98.6\% \\
\quad Center    & 1502.7 & 70.6 & \textbf{63.3} & 87.4 & 98.8\% \\
\quad Farthest  & 1500.3 & \textbf{70.7} & 63.2 & 87.4 & 98.7\% \\
\midrule
\multicolumn{6}{l}{\textit{Retention Ratio = 10\%}} \\
\quad \texttt{[CLS]}  & \textbf{1469.3} & \textbf{70.5} & 59.4 & 85.0 & \textbf{96.0\%} \\
\quad Random    & 1463.7 & 70.2 & 59.4 & 85.0 & 95.8\% \\
\quad Center    & 1455.7 & 70.1 & \textbf{59.5} & 84.8 & 95.6\% \\
\quad Farthest  & 1461.2 & 70.2 & \textbf{59.5} & \textbf{85.1} & 95.8\% \\
\bottomrule
\end{tabular}
\caption{\textbf{Ablations on Pivot Token Selection Strategy.} 
All results are based on LLaVA-NeXT-7B. 
``\texttt{[CLS]}'' = token with highest attention to encoder \texttt{[CLS]}; 
``Center'' = token nearest to mean visual feature; 
``Farthest'' = farthest token from mean; ``Random'' = random token.}
\label{tab:pivot-ablation}
\end{threeparttable}
\end{table}

We conduct an ablation study on different pivot token selection strategies used in the diversity-driven reduction stage. 
As shown in \Cref{tab:pivot-ablation}, selecting the token with the highest attention to the encoder \texttt{[CLS]} token yields the best performance, 
while choosing the token nearest or farthest from the mean visual feature performs less effectively. 
Interestingly, randomly selecting a pivot token achieves comparable performance, suggesting that this strategy can serve as a practical alternative for MLLMs whose encoders do not contain a \texttt{[CLS]} token, thereby making ToDRE more transferable across different model architectures.

\subsection{Layers in Relevance-driven Visual Token Reduction} \label{sec:supp_layer}

\begin{table}[htbp]
    \centering
    \resizebox{\linewidth}{!}{
    \begin{tabular}{l|ccccc|c}
        \toprule[2pt]
        \textbf{Layers} & \textbf{\makecell{Total Time $\downarrow$ \\ (Min:Sec)}} & \textbf{MME} & \textbf{ScienceQA} & \textbf{GQA} & \textbf{POPE} & \textbf{Average} \\
        \midrule
        \rowcolor{gray!25}
        LLaVA-NeXT-7B \cite{liu2024llavanext} & 77:04 & 1519.6 & 72.0 & 64.2 & 87.7 & 100.0\% \\
        \midrule
        $L$ & 78:51 & 1519.6 & 71.8 & 64.2 & 87.6 & 99.9\% \\
        $L/2 \sim L$ & 39:20 & 1527.8 & 65.4 & 39.0 & 87.1 & 87.9\% \\
        $L/2+5L/8+6L/8+7L/8$ & 65:22 & 1527.8 & 65.4 & 39.1 & 87.2 & 87.9\% \\
        $L/2$ & 64:19 & 1527.8 & 65.4 & 39.0 & 87.1 & 87.9\% \\
        $5L/8$ & 58:14 & 1528.6 & 71.8 & 54.3 & 87.5 & 96.2\% \\
        $6L/8$ & 68:54 & 1527.3 & 71.8 & 59.8 & 87.6 & 98.3\% \\
        $\mathbf{7L/8}$ \textbf{(Ours)} & 70:15 & 1522.7 & 71.9 & 64.3 & 87.6 & \textbf{100.0\%} \\
        \bottomrule[2pt]
        \end{tabular}
    }
    \caption{\textbf{Ablation study on selected layers in relevance-driven visual token reduction.} $L$ denotes the total number of decoder layers in the LLM. The bolded row corresponds to the default setting used in the main paper.
}
    \label{tab:ablation_layers}
\end{table}

% We conduct an ablation study to investigate the optimal candidate layer selection when applying relevance-driven visual token reduction strategy (\Cref{sec:method_stage2}) during LLM prefilling.
We conduct an ablation study to investigate the optimal candidate layer selection when applying relevance-driven visual token reduction strategy during LLM prefilling.
As shown in \Cref{tab:ablation_layers}, applying the adaptive reduction at earlier layers (e.g., starting from $L/2$) is suboptimal, 
% as the attention ratios $\alpha_{v \to t}$ and $\alpha_{t \to v}$ (\Cref{eq:alpha})—which characterize the degree of cross-modal interaction—fluctuate considerably in the early layers. 
as the attention ratios $\alpha_{v \to t}$ and $\alpha_{t \to v}$, which characterize the degree of cross-modal interaction, fluctuate considerably in the early layers. 
Early pruning thus prematurely interrupts the ongoing alignment process between modalities. 

In contrast, applying the proposed strategy at later decoder layers (i.e., the last three rows) yields a favorable trade-off between efficiency and performance. 
Although pruning at $5L/8$ or $6L/8$ further reduces the inference time, both settings incur a noticeable drop in average performance (around 2--4\% compared to the full model). 
By contrast, applying the reduction at $7L/8$ restores the performance to the baseline level while maintaining nearly the same computational efficiency. 
% Therefore, the layer selection strategy described in \Cref{sec:method_stage2}—starting from $7L/8$—achieves the best balance between accuracy and efficiency.
Therefore, the layer selection strategy described in the main paper, starting from $7L/8$, achieves the best balance between accuracy and efficiency.

\subsection{Threshold in Relevance-driven Visual Token Reduction} \label{sec:supp_threshold}

\begin{table}[htbp]
    \centering
    \resizebox{\linewidth}{!}{
    \begin{tabular}{l|ccccc|c}
        \toprule[2pt]
        \textbf{Threshold} $\bm{\tau}$ & \textbf{\makecell{Total Time $\downarrow$ \\ (Min:Sec)}} & \textbf{MME} & \textbf{ScienceQA} & \textbf{GQA} & \textbf{POPE} & \textbf{Average} \\
        \midrule
        \rowcolor{gray!25}
        LLaVA-NeXT-7B \cite{liu2024llavanext} & 77:04 & 1519.6 & 72.0 & 64.2 & 87.7 & 100.0\% \\
        \midrule
        0.03 & 79:22 & 1519.6 & 71.7 & 64.2 & 87.6 & 99.9\% \\
        0.05 & 73:24 & 1530.2 & 71.7 & 64.2 & 87.6 & 100.0\% \\
        \textbf{0.10 (Ours)} & 72:35 & 1530.4 & 71.7 & 64.2 & 87.7 & \textbf{100.1\%} \\
        0.15 & 72:25 & 1524.0 & 71.7 & 64.2 & 87.6 & 100.0\% \\
        \bottomrule[2pt]
        \end{tabular}
    }
    \caption{\textbf{Ablation study on threshold $\bm{\tau}$ in relevance-driven visual token reduction.} 
    % When both attention ratios $\alpha_{t \to v}$ and $\alpha_{v \to t}$ (\Cref{eq:alpha}) fall below the threshold $\tau$, all remaining visual tokens are removed from the LLM input.
    When both attention ratios $\alpha_{t \to v}$ and $\alpha_{v \to t}$ fall below the threshold $\tau$, all remaining visual tokens are removed from the LLM input.
}
    \label{tab:ablation_threshold}
\end{table}

We conduct an ablation study on the threshold $\tau$ used in the relevance-driven visual token reduction strategy, as shown in \Cref{tab:ablation_threshold}. 
% Varying $\tau$ controls the aggressiveness of token pruning based on the measured attention ratios $\alpha_{t \to v}$ and $\alpha_{v \to t}$ (\Cref{eq:alpha}).
Varying $\tau$ controls the aggressiveness of token pruning based on the measured attention ratios $\alpha_{t \to v}$ and $\alpha_{v \to t}$.
A larger $\tau$ leads to more extreme pruning but may sacrifice accuracy, while a smaller $\tau$ makes token pruning more conservative and thus improves performance at the cost of increased computational burden.

We choose $\tau = 0.10$ as our default setting, which yields the best trade-off between efficiency and performance: it maintains 100.1\% average accuracy while reducing inference time by over 4 minutes compared to the uncompressed baseline.

\section{Negligible Cross-Attention to Visual Tokens during LLM Decoding} \label{sec:supp_neg}

\begin{figure}[t]
  \centering
  \begin{subfigure}[t]{0.48\linewidth}
    \includegraphics[width=\linewidth]{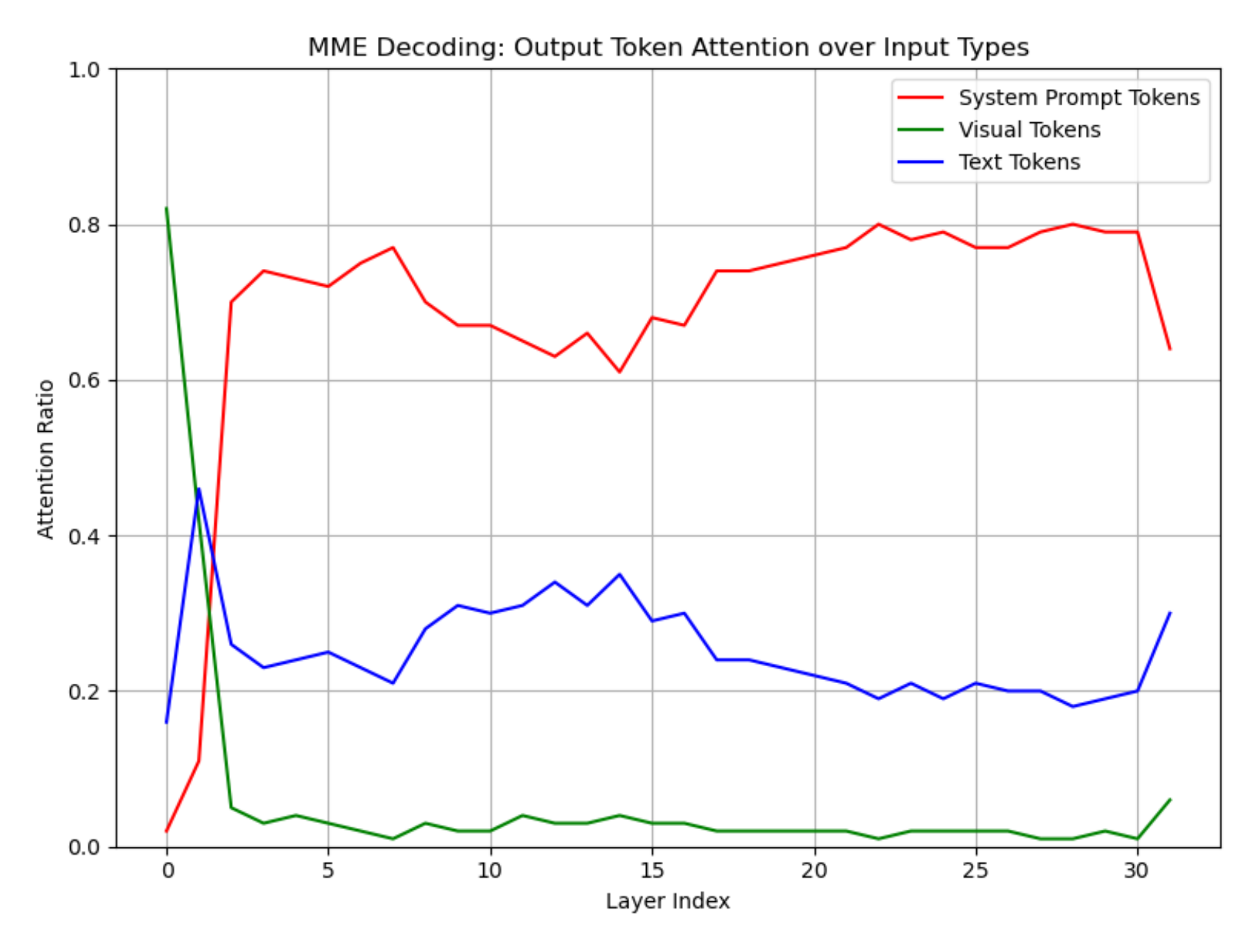}
    \caption{MME Decoding}
  \end{subfigure}
  \hfill
  \begin{subfigure}[t]{0.48\linewidth}
    \includegraphics[width=\linewidth]{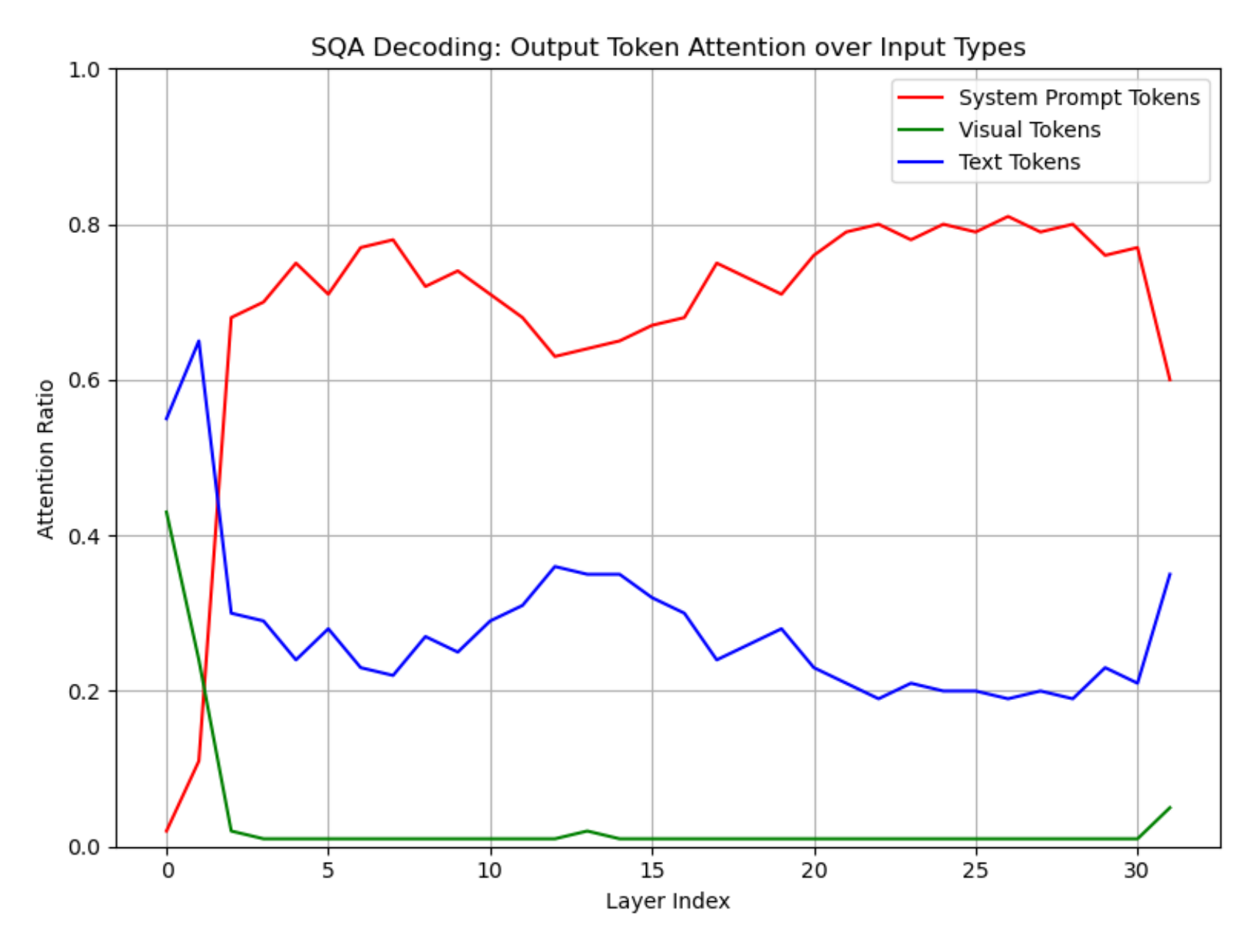}
    \caption{SQA Decoding}
  \end{subfigure}

  \begin{subfigure}[t]{0.48\linewidth}
    \includegraphics[width=\linewidth]{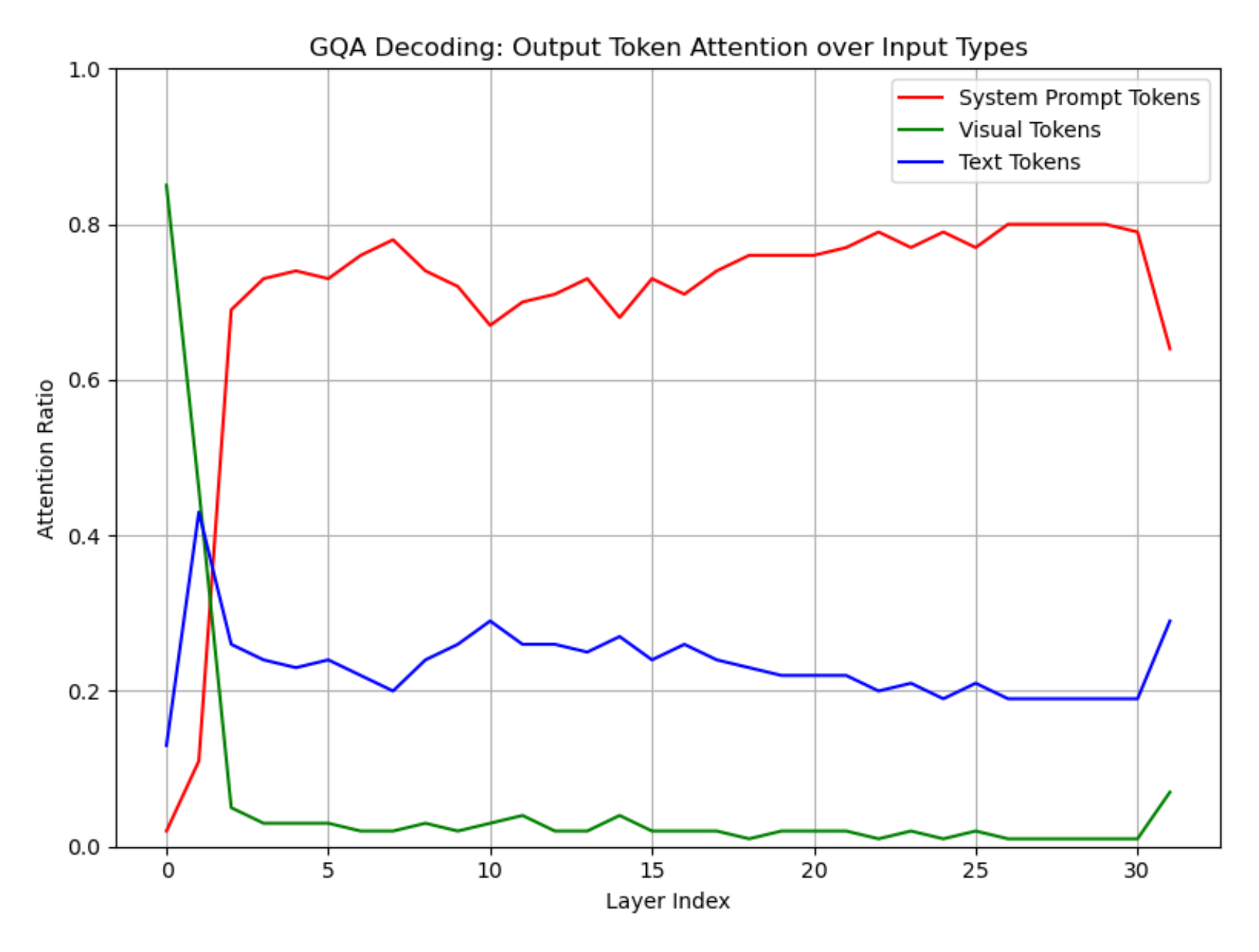}
    \caption{GQA Decoding}
  \end{subfigure}
  \hfill
  \begin{subfigure}[t]{0.48\linewidth}
    \includegraphics[width=\linewidth]{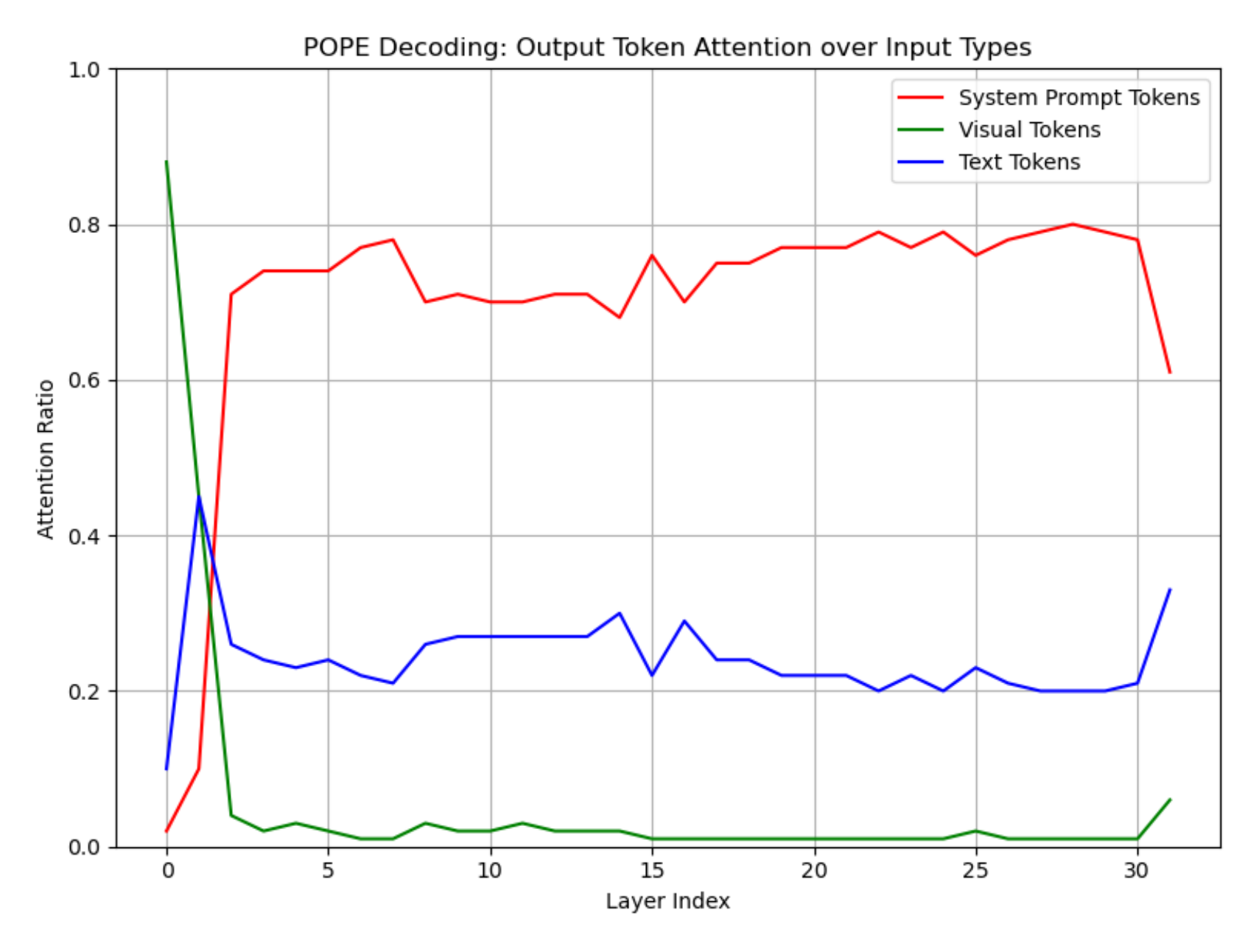}
    \caption{POPE Decoding}
  \end{subfigure}
  
  \caption{Output token attention toward different input token types across LLM layers during decoding. Results are averaged over 100 samples per benchmark.}
  \label{fig:decoding_attn_ratio}
\end{figure}

% Building on the clear \textit{information migration} phenomenon observed during the LLM prefilling stage (\Cref{fig:attn_mig}), we further examine the effect of pruning all visual tokens during \emph{decoding} stage.
Building on the clear \textit{information migration} phenomenon observed during the LLM prefilling stage, we further examine the effect of pruning all visual tokens during the \emph{decoding} stage.
As shown in \Cref{fig:decoding_attn_ratio}, in the shallow layers, attention from output tokens to system prompt tokens increases sharply, while attention directed towards visual tokens drops significantly.
Moreover, in the middle and deeper layers, the output tokens consistently exhibit high attention towards system prompt tokens and text tokens, whereas attention to visual tokens remains negligible (less than 5\%).
These findings further validate the effectiveness of our relevance-driven visual token reduction strategy.

\section{More Case Studies} \label{sec:supp_cases}

\subsection{Free-Form Question Answering with Long Responses} \label{sec:supp_longqa}

\begin{figure}[!t]
  \centering
  \includegraphics[width=\linewidth]{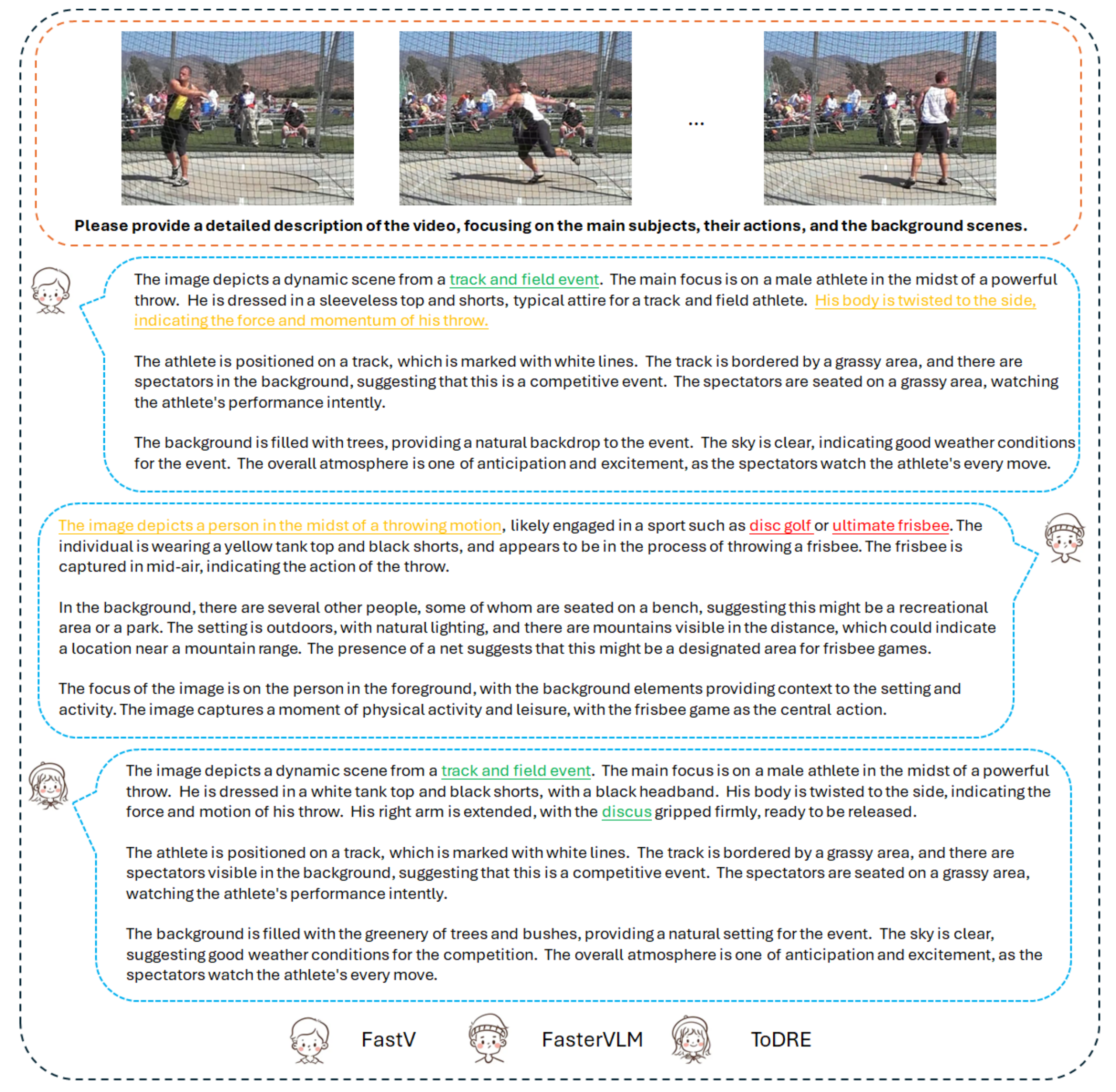}
  \caption{\textbf{Qualitative comparison of free-form video-grounded QA on the Video Detail Caption benchmark \cite{chai2025videodetailcaption}.} \textcolor[rgb]{0,0.69,0.31}{\ul{Green text}} highlights correctly identified events and objects; \textcolor{red}{\ul{red text}} indicates incorrect predictions; \textcolor[rgb]{1,0.75,0}{\ul{yellow text}} marks missing but essential information.}
  \label{fig:supp_long_qa}
\end{figure}

We present qualitative comparisons of free-form question answering with long responses on the Video Detail Caption benchmark \cite{chai2025videodetailcaption}.
As shown in \Cref{fig:supp_long_qa}, our method accurately identifies both the event and activity depicted in the video.
In contrast, FastV \cite{chen2024fastv} produces a vague description of the action and omits key objects, while VisPruner \cite{zhang2024fastervlm} generates a generic caption (``a throwing motion'') and incorrectly identifies the main object.
These comparisons highlight the superior descriptive precision of our approach in capturing fine-grained visual details.

\begin{figure}[!t]
  \centering
  \includegraphics[width=\linewidth]{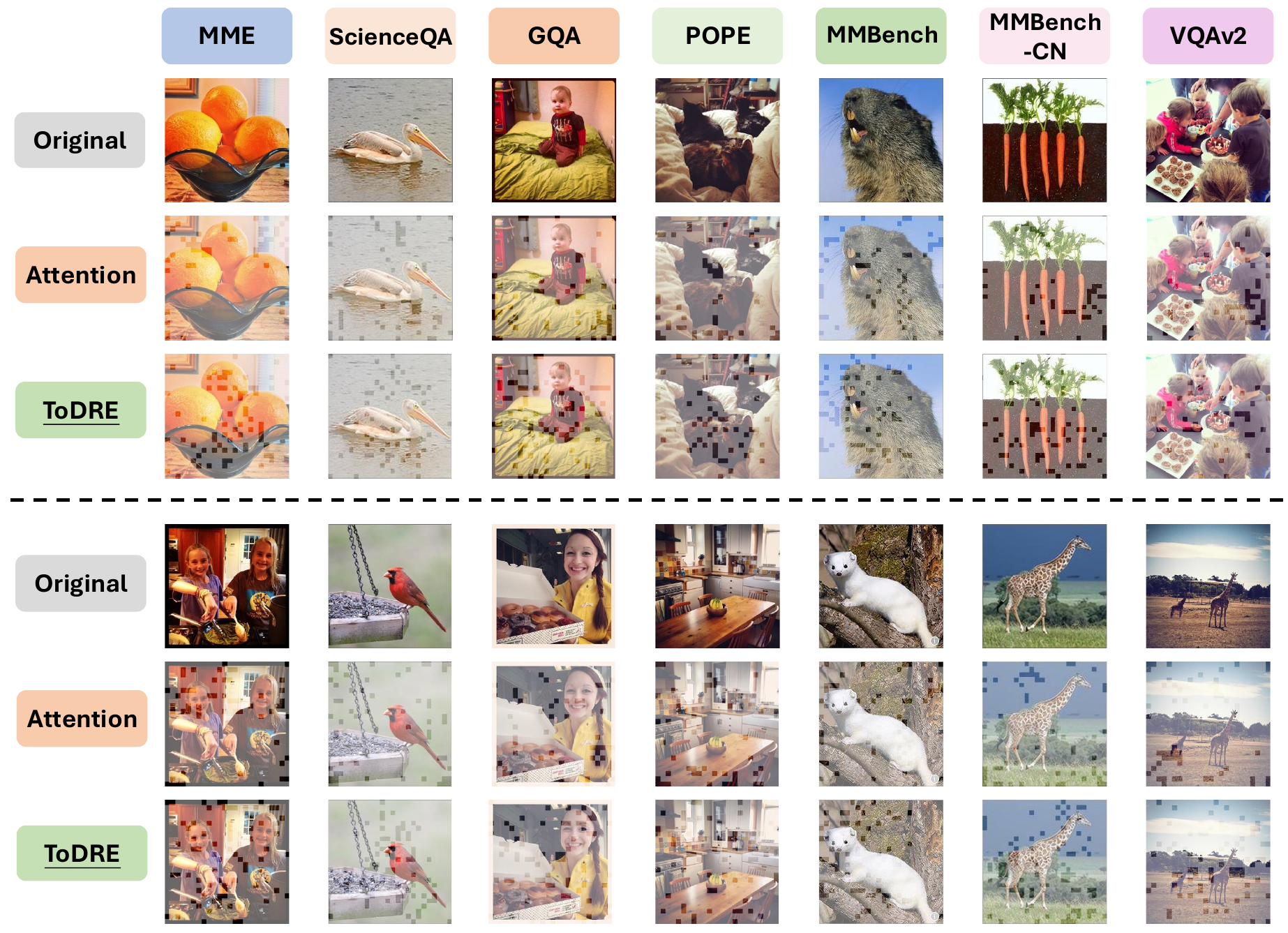}
  \caption{\textbf{Supplementary visualizations comparing attention-driven and ToDRE-based token compression.} The visualization is based on seven benchmarks: MME \cite{Fu2023mme}, SQA \cite{lu2022sqa}, GQA \cite{hudson2019gqa}, POPE \cite{li2023pope}, MMBench and MMBench-CN \cite{Liu2023mmbench}, VQAv2 \cite{goyal2017vqav2}. Best viewed when zoomed in.}
  \label{fig:supp_attn_div}
\end{figure}

\subsection{Attention-Driven Token Pruning vs. ToDRE Token Retention} \label{sec:supp_visualization}

We provide additional visualizations of attention-based token reduction and ToDRE token retention across various image understanding benchmarks, including MME \cite{Fu2023mme}, SQA \cite{lu2022sqa}, GQA \cite{hudson2019gqa}, POPE \cite{li2023pope}, MMBench and MMBench-CN \cite{Liu2023mmbench}, VQAv2 \cite{goyal2017vqav2}.
As shown in \Cref{fig:supp_attn_div}, attention-based token retention tends to produce more concentrated token distributions, focusing on a limited subset of high-attention regions.
In contrast, ToDRE retention results in a more dispersed selection of tokens, covering broader spatial and semantic regions. This broader coverage enables the model to better handle a wider array of open-ended questions.

\end{document}